%% file: main.tex
\documentclass[]{grounding-template}
\input{compile_compat.tex}
\input{layout}
\title{ZoomLynx: Learning to Zoom In on What Matters}
\input{sections/frontmatter}

\newcommand{\listedsection}[1]{%
  \Needspace{3\baselineskip}%
  \section*{\phantomsection #1}%
  \addcontentsline{toc}{section}{#1}}

\begin{document}
\maketitle
\smallskip
\phantomsection
\addcontentsline{toc}{section}{Abstract}
{\centering\sffamily\bfseries Abstract\par}
{\setlength{\baselineskip}{11pt}\noindent\input{sections/abstract}\par}
\par\noindent\begin{minipage}{\linewidth}
\setcounter{tocdepth}{3}
\pdfbookmark[1]{Contents}{zoomlynxcontents}
\tableofcontents
\end{minipage}\par
\setcounter{tocdepth}{3}
\clearpage
\input{sections/body}
\input{sections/statements}

\renewcommand{\bibsection}{\listedsection{\refname}}
\bibliographystyle{ieeenat_fullname}
\bibliography{main}
\addtocontents{toc}{\protect\setcounter{tocdepth}{1}}
\input{sections/appendix}
\end{document}

%% file: layout.tex
\usepackage{amsmath,amssymb}
\usepackage{enumitem}
\usepackage{colortbl,tabularx,wrapfig,needspace}
\setcitestyle{numbers,square,sort&compress}
\newcount\ArxivPreviousLines
\newcount\ArxivWrapSteps
\makeatletter
\newcommand{\FinishArxivWrap}{\par
  \ArxivWrapSteps=0
  \loop\ifnum\c@WF@wrappedlines>1
    \ArxivPreviousLines=\c@WF@wrappedlines
    \noindent\strut\par
    \advance\ArxivWrapSteps by 1
    \ifnum\c@WF@wrappedlines<\ArxivPreviousLines\else\WFclear\fi
    \ifnum\ArxivWrapSteps>80\WFclear\fi
  \repeat\WFclear}
\makeatother
\newtcolorbox{paperprompt}[1][]{
  enhanced, breakable, sharp corners,
  colback=black!3, colframe=black!32, boxrule=0.45pt,
  boxsep=0pt, left=6pt, right=6pt, top=4pt, bottom=4pt,
  before skip=6pt plus 2pt, after skip=6pt plus 2pt,
  fontupper={\fontsize{9}{10.25}\selectfont\ttfamily},
  before upper={\raggedright\setlength{\parindent}{0pt}\setlength{\parskip}{2pt}},
  #1
}

\newtcolorbox{appendixcase}{
  enhanced, sharp corners,
  colback=templatetitle!2!white, colframe=templatetitle!32!white,
  boxrule=0.45pt, boxsep=0pt, left=7pt, right=7pt, top=4pt, bottom=4pt,
  before skip=5pt plus 1pt, after skip=5pt plus 1pt,
  before upper={\setlength{\parindent}{0pt}\setlength{\parskip}{0pt}}
}

%% file: sections/frontmatter.tex
\author[1,*]{Wei~Wei}
\author[1,*]{Dongzi~Shi}
\author[1,2,*]{Qize~Yu}
\author[1]{Xini~Ding}
\author[7]{Hongming~Chen}
\author[6]{Wenxuan~Song}
\author[2]{Minghua He}
\authorbreak
\author[5]{Bohan Zhou}
\author[2]{Jiaqi~Liang}
\author[4]{Yuran~Wang}
\author[2]{Yue~Chen}
\author[1]{Lianrui~Fan}
\author[1]{Hao~Li}
\author[1]{Xianming~Liu}
\author[6]{Haoang~Li}
\author[3,\dagger]{Ping~Luo}
\author[1,\dagger]{Yi~Zou}
\author[1,\dagger]{Shuchang~Zhou}
\affiliation[1]{XPeng Inc.}
\affiliation[2]{Peking University}
\affiliation[3]{The University of Hong Kong}
\affiliation[4]{National University of Singapore}
\affiliation[5]{The Chinese University of Hong Kong}
\affiliation[6]{HKUST (GZ)}
\affiliation[7]{Sun Yat-sen University}
\contribution[*]{Equal contribution}
\contribution[\dagger]{Corresponding authors}
\DeclareRobustCommand{\coverresourcelink}[1]{\href{#1}{\vphantom{#1}{\sffamily \fontsize{8.8pt}{11pt}\selectfont \texttt{#1}}}}
\project{\coverresourcelink{https://skywalker-yqz.github.io/ZoomLynx/}}
\code{\coverresourcelink{https://github.com/Skywalker-yqz/ZoomLynx}}
\model{\coverresourcelink{https://huggingface.co/Skywalker0410/ZoomLynx}}
\evaldata{\coverresourcelink{https://huggingface.co/datasets/Skywalker0410/Aperture-580}}
\titleteaser{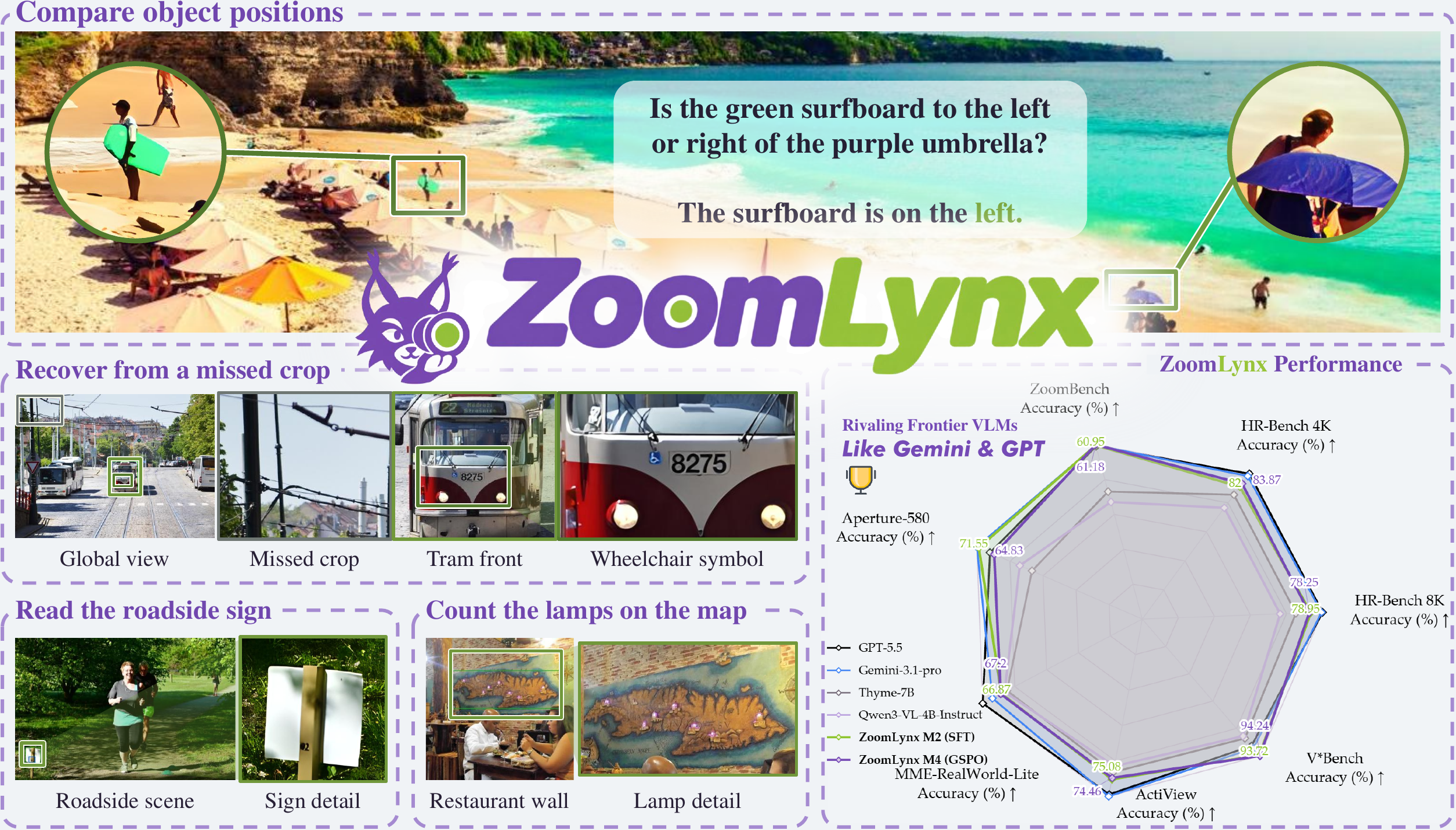}
\abstract{}

\makeatletter
\newlength{\coverfigurewidth}
\renewcommand{\mymaketitle}{%
  \setlength{\coverfigurewidth}{\textwidth}%
  \tcbset{enhanced,frame hidden}%
  \tcbset{left=0.5cm}%
  \tcbset{right=0.5cm}%
  \tcbset{top=0.26cm}%
  \tcbset{bottom=0.26cm}%
  \tcbset{arc=10pt}%
  \tcbset{colback=templatebackground}%
  \tcbset{before skip=0pt}%
  \tcbset{grow to left by=1.5pt}%
  \tcbset{grow to right by=1.5pt}%
  \begin{tcolorbox}
    \hypersetup{linkcolor=templatetitle,citecolor=templatetitle,urlcolor=templatetitle}%
    \setlength{\parindent}{0cm}%
    \setlength{\parskip}{0pt}%
    {\setlength{\parskip}{0cm}%
      \raggedright
      \nohyphens
      {\setstretch{1.618}\titlelist\par}%
      \vskip 0.12cm
      \authorlist\par
      \vskip 0.12cm
      \affiliationlist\par
      \contributionlist\par
    }%
    \ifdefempty{\teaserfile}{}{%
      \smallskip
      {\setlength{\parskip}{0pt}%
        {\centering
          \captionsetup{type=figure}
          \noindent\makebox[\linewidth][c]{\includegraphics[width=1.0\linewidth,page=1]{figures/Teaser-ARXIV.pdf}}\par
          \caption{ZoomLynx zooms in to inspect fine details. It rivals frontier VLMs across seven benchmarks.}
          \label{fig:teaser}
        }%
        \kern-3pt
        \templatedashedrule
        \kern3pt
      }%
    }%
    \ifdefempty{\metadatalist}{}{%
      \vskip 0.04cm
      \noindent
      \begin{minipage}[c]{\linewidth}
        \setlength{\parskip}{0cm}%
        \raggedright
        \metadatalist
      \end{minipage}\par
      \vskip 0.04cm
    }%
  \end{tcolorbox}%
  \tcbset{reset}%
  \enlargethispage{\baselineskip}%
  \FloatBarrier
}

\makeatother

%% file: sections/abstract.tex
Active reasoning matters. High-resolution visual tasks require models to seek local evidence and adapt their next move to what they observe. Yet base vision-language models may repeat uninformative zoom calls without correcting localization or recognizing when to stop. Existing approaches compensate for localization errors through multi-round search, region refinement, and feedback verification, increasing inference cost. Effective visual reasoning therefore calls for a structured observation process in which reasoning guides evidence acquisition and actual observations shape subsequent decisions. We introduce ZoomLynx, a visual reasoning agent named for the lynx’s keen sight and agility. Using a single zoom tool, ZoomLynx organizes evidence acquisition around three explicit decisions: When to inspect, Where to zoom, and when to Stop. To supervise these decisions, we develop a data pipeline that converts real tool interactions into structured training trajectories tailored to the model’s current capabilities. Drawing on both successful and failed observations, the pipeline provides complementary supervision for selective inspection, recovery, and stopping. We learn observation policies from these trajectories through Selective Zoom Training and subsequent preference optimization. We further introduce Aperture-580, a benchmark for a more comprehensive evaluation of active visual reasoning. On ZoomBench and V*Bench, the 4B GSPO model scores above GPT-5.5 and Gemini-3.1-pro in their direct-answer settings. Behavioral analyses show reduced unnecessary calls and improved evidence localization. These results demonstrate the value of coupling structured observation data with policy learning to enable purposeful zooming for active visual reasoning.

%% file: sections/body.tex
\section{Introduction}

Vision-language models (VLMs) use tools such as cropping, magnification, visual search, and code execution to obtain task-relevant visual information during reasoning ~\cite{wu2024vstar,zheng2026deepeyes,zhang2026thyme,hu2024visual}. Zooming, which crops and enlarges an image region, is an observation operation that a model can apply repeatedly to refine a target region and inspect local details ~\cite{wu2024vstar,zheng2026deepeyes}. Each call changes the available evidence, making active observation a sequential decision process: the model decides whether to use a tool and then whether to continue inspecting, answer, or abstain ~\cite{lai2025minio3,zheng2026deepeyes,lin2025adaptvision,li2026aimcot,ma2026act2see,bousselham2026loci}.

\textbf{The challenge is to use a limited number of calls to acquire useful evidence.} When an initial crop misses the required evidence, a base model may continue inspecting irrelevant regions or answer without enough information. Multi-round search, region refinement, and feedback verification can provide additional evidence, but each step requires another tool interaction  ~\cite{lai2025minio3,li2026deepscan,bousselham2026loci}. We therefore ask how training can help a model selectively acquire local evidence, recover from faulty observations, and abstain when it cannot answer reliably within the available budget.

\textbf{We introduce ZoomLynx, an active visual reasoning agent equipped with a single zoom tool.} Figure~\ref{fig:teaser} illustrates the observation policy. We collect trajectories with \textit{the model in the loop}. When a crop misses its target, we keep the policy's call and the image it actually received. A teacher continues from this prefix, and we supervise the corrected continuation. The resulting training data include both successful observations and recoveries from the policy's own mistakes.

The training mixture covers five behaviors. Direct examples teach the policy to answer from the original image. Necessary-Zoom and Multi-region examples teach it to inspect a local detail or combine evidence from several regions. Recovery examples teach it to try another crop when the first misses its target, and Stop/Abstain examples teach it to stop when the available evidence cannot support an answer. We first apply
supervised fine-tuning (SFT) to the base model, then apply direct preference
optimization (DPO) and group sequence policy optimization (GSPO) separately to
the same SFT model and compare the two branches on answer accuracy and
observation behavior.

To distinguish observation behavior from answer accuracy, we build the \textbf{Aperture-580} benchmark and freeze its four task categories before evaluation: direct answering, necessary inspection, multi-region evidence acquisition, and unanswerable questions. Its evidence-region annotations support analyses of tool use, localization, and abstention. We also run paired recovery evaluations under a shared faulty initial crop, in which the base model and its SFT, DPO and GSPO variants continue from the same returned observation on the same question.

The complete SFT recipe improves both answer accuracy and observation behavior. It reduces calls on directly answerable questions, increases necessary-inspection recall, and improves evidence recovery and final answers after a shared faulty initial crop. Our 4B models achieve the best results on most benchmarks among the evaluated open-source models of comparable size. The DPO branch scores 62.60 on ZoomBench and 94.76 on V*Bench, exceeding the Direct results of GPT-5.5 and Gemini-3.1-pro on both benchmarks. Subsequent optimization introduces trade-offs: the GSPO branch reduces some calls while also lowering necessary-inspection recall and abstention on unanswerable questions.

In summary, our major contributions are as follows:
\begin{itemize}
\item \textbf{Preserving real failures and repairing them.} We keep the model's
failed calls and their observations, and a teacher repairs each trajectory so the
model learns from its own errors rather than having them filtered out.
\item \textbf{Training a budgeted observation policy.} ZoomLynx reduces observation to three decisions: whether to inspect, where to look, and when to answer or abstain. We train this policy with SFT, then compare two post-training branches on answer accuracy and observation behavior.
\item \textbf{Building a diagnostic benchmark.} We built Aperture-580 with annotated task categories and evidence regions, enabling analysis of model
behavior beyond final-answer accuracy.
\end{itemize}

\section{Related Work}
\label{sec:related-work}

\subsection{Visual Tool-Augmented Reasoning}

External visual tools extend the perception and reasoning capabilities of VLMs. Visual Sketchpad incorporates visual operations as intermediate steps in the reasoning process ~\cite{hu2024visual}, while DeepEyes, Pixel Reasoner, and VTool-R1 interleave textual reasoning with visual operations, enabling iterative observations through multiple tool interactions ~\cite{zheng2026deepeyes,su2025pixel,wu2026vtool}. Beyond tool integration, training strategies also play a central role in improving tool-use capabilities. T3-Agent constructs and verifies tool-use trajectories for supervised training ~\cite{gao2025t3agent}. Different training paradigms have also been explored~\cite{ping2026longact}: DeepEyes develops visual tool-use capabilities through end-to-end reinforcement learning without cold-start SFT ~\cite{zheng2026deepeyes}, Pixel Reasoner adopts a two-stage pipeline combining instruction tuning with reinforcement learning ~\cite{su2025pixel}, and VTool-R1 employs reinforcement learning with outcome-based rewards to optimize multimodal tool-use behavior ~\cite{wu2026vtool}. ToolsRL introduces direct tool supervision through tool-specific rewards before optimizing task accuracy ~\cite{dong2026toolsrl}, while Learning to Select Visual Tools from Experience uses reinforcement learning to discover query-specific tool selection strategies ~\cite{huang2026vista}. Thyme further broadens the range of available operations through code-based image processing ~\cite{zhang2026thyme}.

\subsection{Zoom-Based Active Visual Reasoning}

For fine-grained understanding of high-resolution images, V* uses visual search to locate question-relevant regions ~\cite{wu2024vstar}. Mini-o3 constructs cold-start trajectories that include trial and error and self-reflection, then extends multi-round search through subsequent training ~\cite{lai2025minio3}. DeepScan studies training-free visual search ~\cite{li2026deepscan}, while LOCI employs a Locator-Critic refinement loop to iteratively improve visual localization and evidence verification ~\cite{bousselham2026loci}. VisTRA further analyzes tool-output verification in small-object visual question answering, showing the importance of checking returned visual evidence rather than accepting tool outputs directly ~\cite{sugiyama2025vistra}. AdaptVision acquires local visual information according to task requirements to balance accuracy and computational cost ~\cite{lin2025adaptvision}. Zooming without Zooming uses region-to-image distillation to improve fine-grained perception in a full-image model ~\cite{wei2026zooming}. Thinking with Visual Grounding further supervises intermediate reasoning with explicit visual evidence regions ~\cite{zhang2026thinkingvg}.

Despite these advances in visual search, evidence acquisition~\cite{yu2026groundingpi,yu2026groundanything}, and verification, learning from ineffective observations remains an important challenge. Prior work has studied silent tool-error detection and recovery in general tool-use settings ~\cite{sun2024toolsfail}. In contrast, we focus on visual observation failures and use failure-driven continuation supervision to train models to select tool calls, recover from ineffective observations, and abstain when further inspection is unlikely to help under a fixed interaction budget.

\section{Method}
\subsection{Budgeted Observation}

High-resolution questions often depend on a small region that the full image cannot resolve, and indiscriminate zooms waste the limited budget on irrelevant areas. We therefore cast inference as sequential decision-making under a tool budget: each action conditions on the current evidence and determines what is visible next, until the model answers or declares the evidence insufficient.

\paragraph{Notation.} Let $I_0$ be the original image, $q$ the question, $h_{t-1}$ the tool interactions collected before step $t$, and $b_t$ the remaining budget. The policy $\pi_\theta$ produces the action at step $t$:
\begin{center}
$a_t=\pi_\theta(I_0,q,h_{t-1},b_t).$
\end{center}

\paragraph{Budget.} The budget counts the tool calls still available. It starts at $b_1=B=3$. Each call to $\operatorname{zoom}$ reduces it by one, while $\operatorname{answer}$ and $\operatorname{stop}$ leave it unchanged, so $\operatorname{zoom}$ is unavailable once $b_t=0$ and only terminal actions remain.

\paragraph{Action space.} The action space is
\begin{center}
$\mathcal A_t=\left\{\operatorname{answer}(y),\operatorname{stop}(s)\right\}\cup\left\{\operatorname{zoom}(j,r)\mid b_t\geqslant1\right\}.$
\end{center}
The action $\operatorname{zoom}(j,r)$ requests a local observation of image $j$ from the current list of visible images. The region $r=[x_1,y_1,x_2,y_2]$ uses normalized $xyxy$ coordinates in $[0,1000]$ relative to that image. The action $\operatorname{answer}(y)$ returns the final answer $y$, and $\operatorname{stop}(s)$ reports that the available evidence does not support a reliable answer, together with a reason.

\paragraph{Tool and observations.} The model uses one external tool, \texttt{zoom\_and\_super\_resolve(image\_idx, bbox\_2d)}. Executing $\operatorname{zoom}(j,r)$ calls this tool with image index $j$ and region $r$, and the executor returns a new observation:
\begin{center}
$o_t=\operatorname{Executor}(I_0,a_t),$
\end{center}
which contains a cropped image (Raw Crop), an optional super-resolved image (SR Crop), and status text. The Raw Crop carries the factual evidence, and the SR Crop serves as an auxiliary view. The executor sets the magnification factor from the crop size, and the action interface exposes no parameter for choosing the super-resolution algorithm. Because super-resolution can introduce details that the original pixels do not support, a high-confidence answer cannot rest on the enhanced view alone.

\paragraph{History.} The history records earlier actions together with the observations they returned:
\begin{center}
$h_{t-1}=\{(a_1,o_1),\ldots,(a_{t-1},o_{t-1})\},$
\end{center}
where $o_i$ is the observation returned after executing $a_i$. After each step the new pair is appended in chronological order:
\begin{center}
$h_t=h_{t-1}\cup\{(a_t,o_t)\},$
\end{center}
where $\cup$ denotes appending rather than an unordered union. If $a_t$ is $\operatorname{answer}$ or $\operatorname{stop}$, the interaction ends without a new observation.

\subsection{When, Where, and Stop}
The three dimensions decide whether more evidence is needed, which evidence to obtain, and whether observation should end. The last decision yields $\operatorname{answer}$ when the evidence is sufficient and $\operatorname{stop}$ when it is not, regardless of how many calls remain. The model treats a missing target as insufficient evidence and stops, since no later call can find it. In a counting question, the same absence instead yields an answer of zero. Recovery denotes obtaining the missing evidence through later actions and completing the answer after an ineffective or insufficient observation.

\subsection{Data Construction}
\label{sec:data}

\begin{figure}[!htbp]
  \centering
  \includegraphics[width=1.0\linewidth]{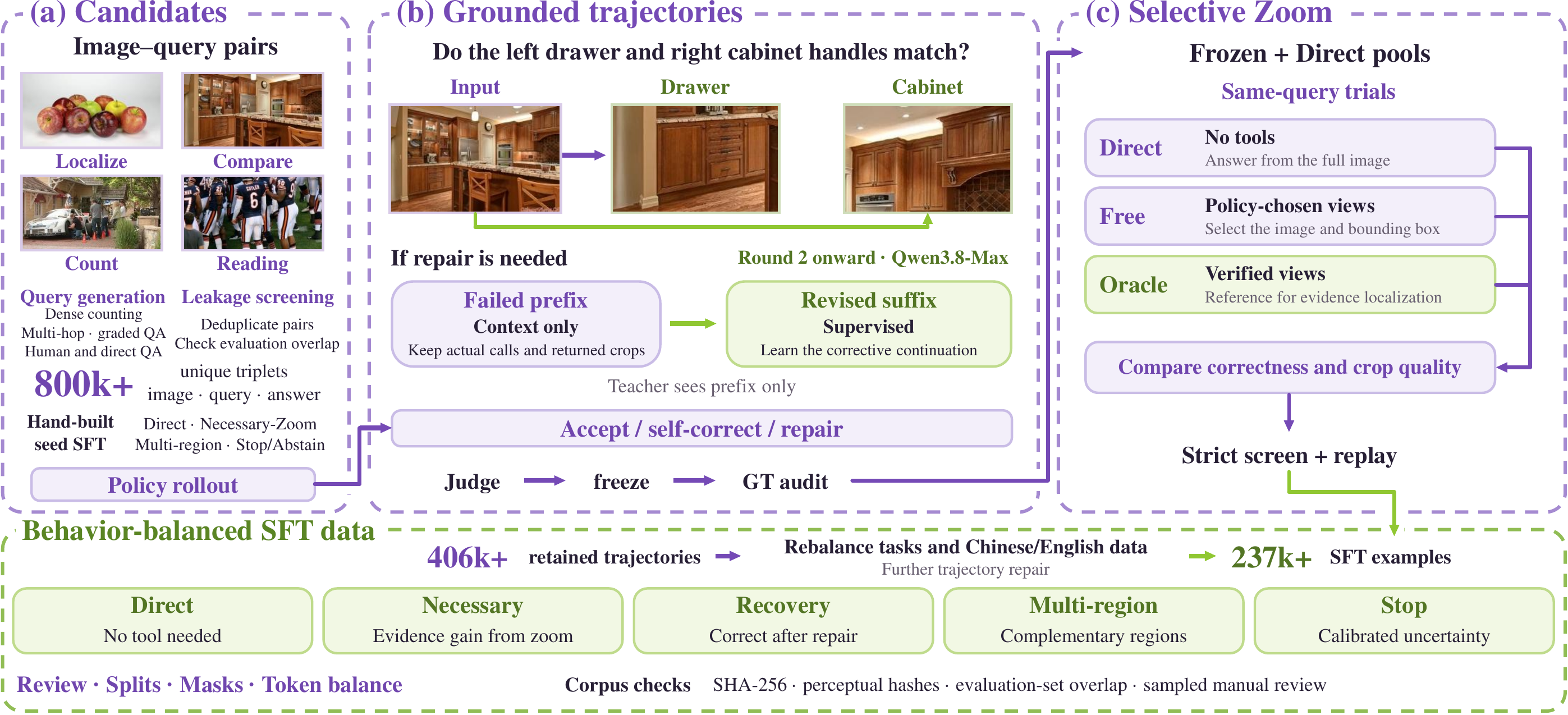}
  \caption{\textbf{Selective zoom data construction.} Candidate questions (a) yield tool-use trajectories and teacher repairs (b). Direct, Free, and Oracle trials (c) guide selection across five observation behaviors.}
  \label{fig:data_pipe}
\end{figure}

Our candidate pool contains roughly 800k unique image--query--answer triplets drawn from public datasets and human-written questions, including dense counting, multi-hop, and graded questions (Figure~\ref{fig:data_pipe}a, with examples in Figure~\ref{fig:qa-construction}). We screen candidates for overlap with evaluation images before rollout.

\subsubsection{Model-in-the-Loop Trajectory Construction}

We train a seed policy on a small hand-built set without Recovery examples. In each round, the current policy calls the zoom tool on new questions (Figure~\ref{fig:data_pipe}b). From round two onward, Qwen3.8-Max can continue a rollout after a failed first crop. The teacher sees only the recorded prefix, including the policy's calls and the images returned. We keep the prefix as context and supervise the corrected continuation (see Figure~\ref{fig:repair-trajectory-supp}).

We compare three answers to the same question, using no tools (Direct), policy-chosen views (Free), or verified regions (Oracle). Answer correctness and crop quality determine which trajectories advance to replay (Figure~\ref{fig:data_pipe}c). Reviewers sample by failure mode, check the recorded calls and crops, and verify that the evidence supports the answer~\cite{huang2026memsteer}. They reject errors unless the evidence supports a unique correction.

After judging, freezing, and a ground-truth audit, we retain 406k+ trajectories. Before inclusion, we deduplicate source images and check evaluation overlap with SHA-256 and perceptual hashes. Further repair and balancing across tasks and languages (Chinese and English) yield 237k+ SFT examples. We balance behaviors by token count so that long trajectories do not dominate training. The five behaviors in Figure~\ref{fig:data_pipe} can overlap. A comparison requiring two regions, as in Figure~\ref{fig:multi-region}, also teaches recovery when the policy initially crops the wrong one.

\subsection{Training Setup}

\begin{figure}[!htbp]
  \centering
  \includegraphics[width=1.0\linewidth]{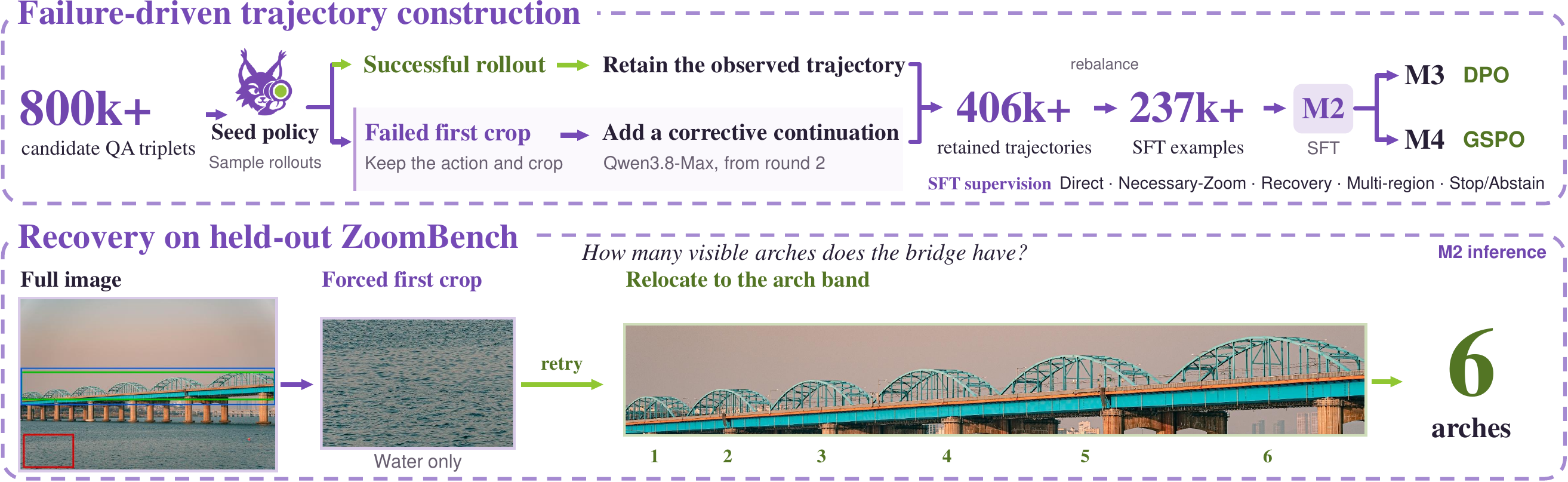}
  \caption{\textbf{Training and recovery at inference.} DPO (M3) and GSPO (M4) start from the same SFT policy (M2). After a forced crop shows only water, M2 locates the bridge and counts six arches.}
  \label{fig:method}
\end{figure}

We use the mixture from Figure~\ref{fig:data_pipe} to train the SFT policy (M2). Figure~\ref{fig:method} shows M2 finding the bridge after a forced first crop shows only water. We fine-tune Qwen3-VL-4B-Instruct for 3 epochs on 16 H800 GPUs across two nodes, with a global batch size of 32, AdamW at a learning rate of $5\times10^{-6}$, a maximum sequence length of 16,384 tokens, and up to 1,024 image tokens per image; training uses bf16 and FlashAttention-2.

Starting from M2, we obtain M3 and M4 with DPO and GSPO, respectively. For DPO, we collect M2 rollouts on the SFT data and keep incorrectly answered examples to build two preference sets: at the answer level, the teacher's correct answer is preferred over the policy's incorrect one, yielding 9k+ category-balanced pairs across position, OCR, counting, and color tasks; at the decision level, we prefer Zoom-before-answering when observation is necessary and direct answering otherwise, yielding 7k+ pairs. Together they contain 17k+ pairs. Images are resized to at most 2 MP and the multimodal context is limited to 8,192 tokens. We train LoRA adapters on M2 (rank 16, $\alpha=32$, dropout 0.05) for 1 epoch.

For GSPO, we train LoRA adapters on M2 with 4.9k+ examples from manual annotations and SFT data, using rewards for answer correctness, protocol compliance, and tool-call cost. We sample $K=8$ responses per group with a batch size of 16, filter groups by within-group reward variance, and allow up to 6 generation batches. Training runs on 8 GPUs, with vLLM generating rollouts at tensor parallelism 4, temperature 0.8, and top-$p$ 0.95. Super-resolution is served by a separate Real-ESRGAN service.

\subsection{Aperture-580}
\label{sec:aperture}

ZoomBench~\cite{wei2026zooming}, HR-Bench~\cite{wang2025hrbench}, V*Bench~\cite{wu2024vstar}, TreeBench~\cite{wang2025traceable}, MME-RealWorld-Lite~\cite{zhang2025mmerealworld}, and ActiView~\cite{wang2025actiview} evaluate high-resolution visual reasoning primarily through answer accuracy.

The same accuracy can hide unnecessary calls, missed observations, or poorly localized crops. Aperture-580 pairs each question with a task bucket and an oracle region to measure over-call rate, necessary-observation recall, localization, and abstention. We test recovery separately with a fixed faulty first crop.

\FloatBarrier

\input{sections/experiments}

\section{Conclusion}

We presented ZoomLynx, a budgeted observation policy that decides whether to
inspect, where to look, and when to answer or abstain. Rather than discarding
failed attempts, its training corpus keeps the model's failed calls and their
observations and has a teacher repair each trajectory, so the policy learns from
its own errors. We also built Aperture-580, which annotates task categories and
evidence regions, enabling diagnosis of observation behavior beyond final-answer
accuracy. Across benchmarks, this training improves answer accuracy and
observation behavior, including selective tool use, recovery from a bad first
crop, and abstention when evidence is insufficient. Future work will evaluate view selection across longer sequences
and explore the use of VLM agents for decision-making in embodied systems.

%% file: sections/experiments.tex
\section{Experiments}
\label{sec:experiments}

\subsection{Benchmark Results}

\subsubsection{Comparison with Open-Source Methods}

\noindent\begin{minipage}[t]{\dimexpr0.35\textwidth-\columnsep\relax}
\vspace{0pt}\raggedright
Figure~\ref{fig:radial} compares M2 and M4 with the leading baselines. We analyze tool use and localization on ZoomBench and Aperture-580.

\medskip
\textbf{SFT accounts for most of the gains.} M2 gains 13.85 points over M1 on ZoomBench, 4.93 on HR-Bench-4K, 13.62 on V*Bench, and 10.69 on Aperture-580 (Table~\ref{tab:benchmark-results}).

\medskip
Among our variants, M3 leads on ZoomBench (62.60) and V*Bench (94.76). M4 leads on HR-Bench-4K (83.87) and MME-RealWorld-Lite (67.20).

\medskip
Among open-source models of comparable size, the ZoomLynx 4B variants are competitive on most benchmarks (Table~\ref{tab:benchmark-results}). Against 4B ZwZ, M4 gains 5.44 points on ZoomBench, 1.57 on V*Bench, and 0.10 on MME-RealWorld-Lite, and trails by 1.25 on HR-Bench-8K.

\medskip
M4 also scores above GPT-5.5 and Gemini-3.1-pro on ZoomBench and V*Bench in their direct-answer settings.
\end{minipage}\hfill%
\begin{minipage}[t]{0.65\textwidth}
  \vspace{0pt}\centering
  \captionsetup{type=figure,justification=raggedright,singlelinecheck=false}
  \includegraphics[width=\linewidth]{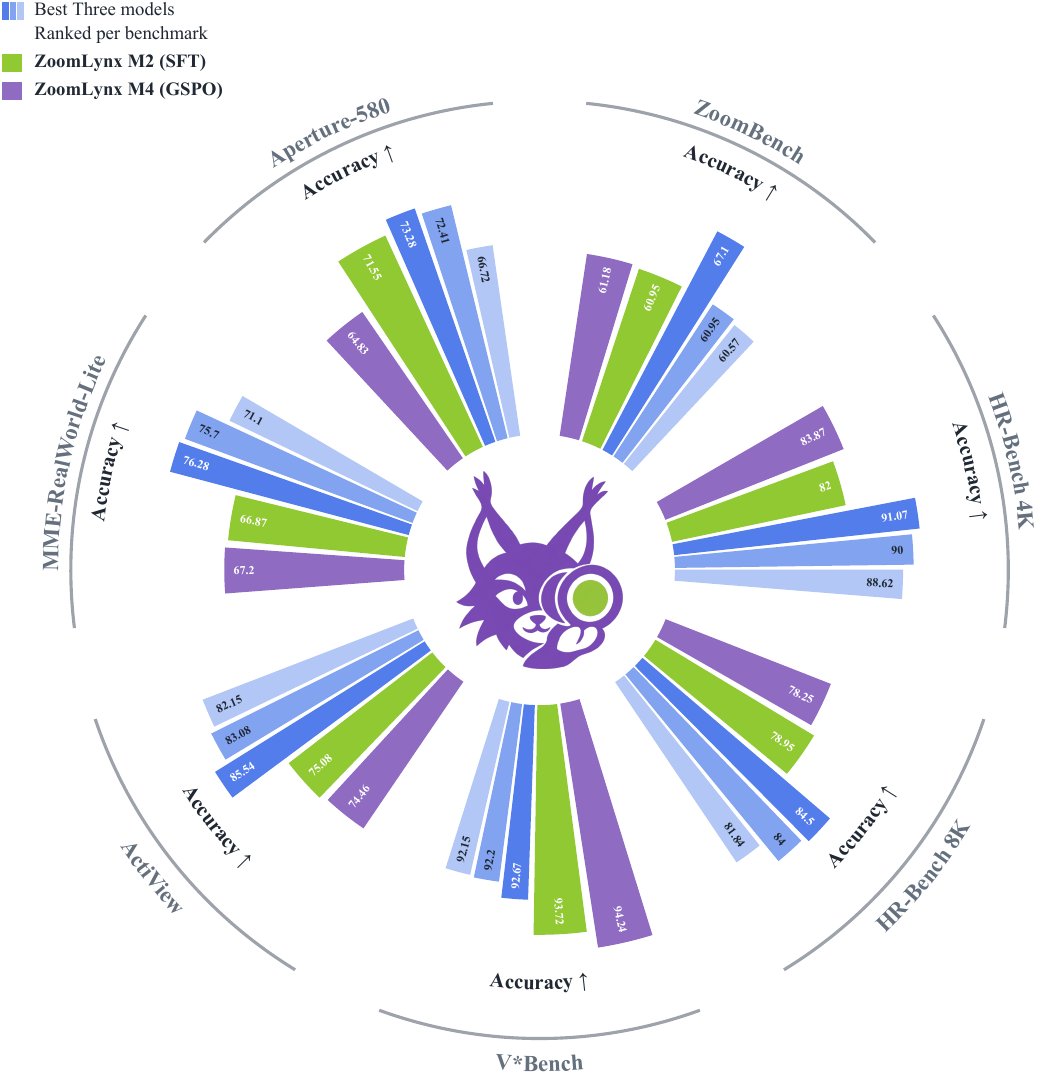}
  \caption{\textbf{Accuracy across seven benchmarks.} ZoomLynx M2 and M4 are compared with the three highest-scoring baselines on each benchmark.}
  \label{fig:radial}
\end{minipage}

\input{tables/benchmark-results}
\FloatBarrier

\subsubsection{Evidence Localization}

We measure localization by first-crop coverage: \emph{contain} (the crop fully covers the oracle region), \emph{tight} (contain with crop-to-oracle area ratio $R\le 2$), and single-call accuracy (answer accuracy when the model makes exactly one tool call). Both coverage metrics require full oracle coverage, so a correctly located but undersized crop scores zero.

\input{tables/localization}

\textbf{Training improves both coverage and single-call accuracy.} From M1 to M3, contain rises from 0.191 to 0.347 on ZoomBench and from 0.474 to 0.716 on Aperture-580, and single-call accuracy from 0.530 to 0.663 and from 0.596 to 0.700 (Table~\ref{tab:localization}). Tight rises much less (0.159 to 0.175; 0.103 to 0.220), because ZoomBench oracles include surrounding context while Aperture-580 uses tight manual boxes.

The gain is therefore not geometric: M2 and M3 are nearly identical, so DPO's increment more likely comes from answer generation and tool-call decisions than from localization. M4 is worse than M3 (contain $-4.0$ and $-4.9$ points, tight $-3.5$ on both, single-call accuracy $-1.3$ and $-2.8$), possibly because the GSPO cost term discourages additional crops.

\subsubsection{Selective Observation, Recovery, and Abstention}

We examine two decision points around a tool call: whether to call when evidence
is needed, and whether to repair an uninformative observation after the tool
returns.

Before a call, \textit{Over-Call Rate} (Direct subset) counts directly answerable examples
that still receive a call, and Necessary-Recall (Necessary subset) counts examples
that receive a call; better values on both indicate more selective tool use.

ZoomBench has no Direct/Necessary labels and provides an official crop region, so
we treat every example as \textit{Necessary-Zoom} and \textit{Necessary-Recall} reduces to the call
rate. A final call is ineffective if it repeats an earlier region or adds no new
image or super-resolution detail; GPT-5.5 resolves uncertain cases. Over-Call Rate
is the fraction of examples with at least two calls whose last call is
ineffective. Because the two benchmarks use different populations, we compare it
only within a benchmark.

\input{tables/call-decisions}

\textbf{SFT improves both necessary observation and call decisions}
(Table~\ref{tab:call-decisions}). On Aperture-580, Over-Call Rate on Direct
examples falls from 0.753 to 0.284 while Necessary-Recall rises from 0.832 to
0.871; on ZoomBench, the call rate rises from 0.557 to 0.813 while Over-Call Rate
falls from 0.273 to 0.111. SFT therefore calls less on directly answerable
questions and obtains local evidence more often when it is needed.

\textbf{DPO reduces unnecessary calls without losing recall, whereas GSPO trades
recall for fewer calls.} From M2, M3 lowers ZoomBench Over-Call Rate from 0.111 to
0.087 with Necessary-Recall essentially unchanged, and raises Aperture-580
Necessary-Recall to 0.884 at an unchanged Over-Call Rate. M4 lowers Direct
Over-Call Rate to 0.204 but drops Necessary-Recall to 0.831; because it changes
both training data and objective, the call-cost term alone cannot explain this
trade-off.

\textit{After a tool returns, the model must answer, retry, or stop.} To fix the
first error, we randomly sample 150 questions per benchmark and give each a bad
first crop with oracle coverage below 0.2; M1--M4 receive the same questions,
first crops, and returned observations, so every comparison is paired (see Figure~\ref{fig:recovery-cases} for examples). Retry requires at
least one later call. Both recovery rates require a correct final answer: absolute
recovery requires a later crop with $\mathrm{cov}\ge0.5$, while
coverage-improvement recovery requires the later crop to raise coverage over the bad crop by at least 0.3. All rates use the 150 controlled examples as the denominator.

\input{tables/controlled-recovery}

\textbf{All three trained models exceed M1 in retry, relocalization, and absolute recovery on both benchmarks} (Table~\ref{tab:controlled-recovery}). M3 improves relocalization by 6.0 percentage points on ZoomBench and 20.0 percentage points on Aperture-580. The localization gain partly reaches the answers. On ZoomBench, absolute recovery for M2, M3, and M4 exceeds M1 by 7.3, 6.6, and 7.3 percentage points, and on Aperture-580 by 15.3, 16.6, and 16.6.

\textbf{Coverage-improvement recovery shows the same pattern, with smaller gains on ZoomBench.} All three trained models gain between 10.7 and 18.7 percentage points over M1, and all six paired differences have 95\% confidence intervals that exclude zero. ZoomBench's official oracles are large, so a fixed 0.5 threshold understates repair there.

\textit{The Stop bucket covers questions that no observation can answer.} Targets may be too small, or images too blurry, for even a high-resolution crop. Aperture-580 includes 74 Stop examples, each manually verified to support no definite answer; we distinguish appropriate abstention, fabrication, and repetitive output.

\input{tables/abstention}

\textbf{SFT improves abstention on unanswerable examples, and the DPO branch retains this gain} (Table~\ref{tab:abstention}). M1 already abstains on some unanswerable questions but still fabricates answers and repeats output. M2 raises appropriate abstention from 70.27\% to 79.73\% and eliminates repetitive output; M3 matches this, whereas M4 falls to 59.45\%. M4 trains without Stop examples, so missing stopping supervision may explain the decline. Fewer calls are therefore not evidence of better observation unless Necessary-Recall and abstention hold.

\subsection{Zero-Shot Extension to Multi-Image and Multi-View Inputs}

ZoomLynx trains on single-image tasks. Because the tool accepts an image index, the same policy can inspect several views at test time without additional training.

\par\noindent\begin{minipage}{\linewidth}
  \centering
  \captionsetup{type=figure}
  \includegraphics[width=1.0\linewidth]{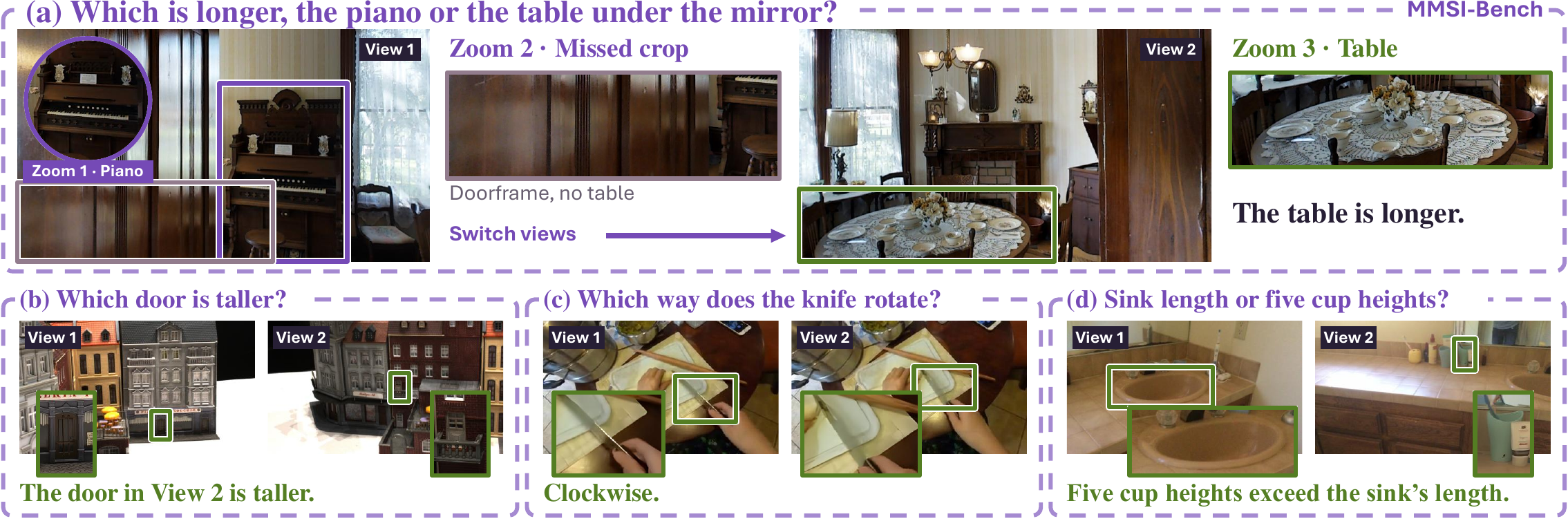}
  \caption{\textbf{Zero-shot reasoning across views.} ZoomLynx switches views after a missed crop on MMSI-Bench. The other examples compare height, rotation, and scale.}
  \label{fig:multi_image}
\end{minipage}\par

In Figure~\ref{fig:multi_image}a, the first crop shows the piano, but the second lands on a doorframe. ZoomLynx switches to the other view, locates the table in the third crop, and correctly judges that the table is longer than the piano.

In panels~b--d, ZoomLynx compares door heights across images, infers the knife's rotation from its orientation in each view, and uses the cup as a scale for the sink. Figure~\ref{fig:multi-view} in Appendix~\ref{app:multi-view} shows the crop sequence for each case. To measure how reliably the policy selects a useful view beyond these four examples, a benchmark could annotate the same object across images and score view selection and object localization separately from the final answer. For video, an analogous test would ask whether a selected frame reveals evidence unavailable in the frames already inspected.

The distinction also matters for training. In Figure~\ref{fig:recovery-failure-stop}, two models answer the sign question correctly despite missing the sign, while all four locate the umbrella but misread its color. Answer-level reward misses the first error, and region overlap misses the second. A step-level reward could instead credit a crop for the evidence it adds~\cite{ping2026longr}. Replaying the same decision with and without that crop could help estimate its contribution. Questions on which the policy overlooks evidence or abstains prematurely would be candidates for further training, especially when finding the evidence takes several observations and the resulting trajectories are harder for a teacher to verify.

%% file: tables/benchmark-results.tex
\begin{table}[!htbp]
\centering
\small
\setlength{\tabcolsep}{2pt}
\renewcommand{\arraystretch}{1.12}

\caption{\textbf{Benchmark accuracy.} Accuracy (\%) is reported for ZoomBench~\cite{wei2026zooming}, HR-Bench-4K and HR-Bench-8K~\cite{wang2025hrbench}, V*Bench~\cite{wu2024vstar}, ActiView~\cite{wang2025actiview}, MME-RealWorld-Lite~\cite{zhang2025mmerealworld}, and Aperture-580. A dagger ($\dagger$) marks baselines with results from their papers or official repositories. GC, CF, RC, TG, and EC denote agents using grounding and cropping, cropping and frame selection, recursive cropping, tree-based grounding, and executable code. TF denotes training-free inference, and 1F denotes a single forward pass. Qwen3-VL-4B with its native crop tool is a reference setting whose Aperture-580 results contain 50\% Stop examples and are not fully comparable with the other rows.}
\label{tab:benchmark-results}

\begin{tabular*}{\textwidth}{
@{\extracolsep{\fill}}
lccrrrrrrr
@{}
}
\toprule
\multirow{2}{*}{\textbf{Model}} & \multirow{2}{*}{\textbf{Size}} & \multirow{2}{*}{\textbf{Setting}}
& \multirow{2}{*}{\footnotesize\textbf{ZoomBench}}
& \multicolumn{2}{c}{\footnotesize\textbf{HR-Bench}}
& \multirow{2}{*}{\footnotesize\textbf{V*Bench}}
& \multirow{2}{*}{\footnotesize\textbf{ActiView}}
& \multirow{2}{*}{\footnotesize\shortstack{\textbf{MME-}\\\textbf{RealWorld-Lite}}}
& \multirow{2}{*}{\footnotesize\shortstack{\textbf{Aperture-}\\\textbf{580}}} \\
\cmidrule(lr){5-6}
& & & & {\footnotesize\textbf{4K}} & {\footnotesize\textbf{8K}} & & & & \\
\midrule

GPT-5.5 (closed)~\cite{openai2026gpt55} & -- & Direct & 60.57 & 90.00 & 84.50 & 87.43 & 82.15 & 75.70 & 66.72 \\
Gemini-3.1-pro (closed)~\cite{google2026gemini31}  & -- & Direct & 60.95 & 88.62 & 84.00 & 89.01 & 83.08 & 71.10 & 72.41 \\
Qwen3.8-max (closed)~\cite{qwen2026qwen38} & -- & Direct & 67.10 & 91.07 & 81.84 & 92.15 & 85.54 & 76.28 & 73.28 \\

\midrule
DeepEyes-7B~\cite{zheng2026deepeyes} $\dagger$ & 7B & GC & 43.55 & 75.10 & 72.60 & 90.10 & 68.31 & 63.50 & 54.31 \\
PixelReasoner-7B~\cite{su2025pixel} $\dagger$ & 7B & CF & 46.27 & 73.38 & 64.69 & 84.30 & 72.62 & 63.50 & 52.41 \\
Mini-o3-7B~\cite{lai2025minio3} $\dagger$ & 7B & RC & 48.17 & 77.50 & 73.30 & 88.20 & 60.62 & 54.90 & 51.90 \\
TreeVGR-7B~\cite{wang2025traceable} $\dagger$ & 7B & TG & 44.85 & 77.10 & 73.10 & 91.10 & 66.15 & 62.90 & 49.14 \\
Thyme-7B~\cite{zhang2026thyme}  $\dagger$ & 7B & EC & 44.73 & 77.00 & 72.00 & 82.20 & 71.69 & 64.20 & 48.28 \\
DeepScan~\cite{li2026deepscan} $\dagger$ & 8B & TF & 50.30 & 73.75 & 71.19 & 92.20 & 72.31 & 61.36 & 51.90 \\
ZwZ-4B~\cite{wei2026zooming}  $\dagger$ & 4B & 1F & 55.74 & 81.75 & 79.50 & 92.67 & 71.08 & 67.10 & 53.97 \\

\midrule
Qwen3-VL-4B-Instruct~\cite{bai2025qwen3vl}  & 4B & Direct & 41.07 & 68.88 & 64.58 & 80.63 & 68.94 & 65.37 & 53.62 \\
Qwen3-VL-4B-Instruct (M1)~\cite{bai2025qwen3vl}  & 4B & Agent & 47.10 & 77.07 & 72.91 & 80.10 & 67.38 & 60.21 & 60.86 \\
Qwen3-VL-4B (native crop)~\cite{bai2025qwen3vl}  & 4B & Agent & 53.37 & 71.13 & 61.86 & 71.73 & 56.90 & 67.08 & 64.83 \\
Qwen3-VL-8B-Instruct~\cite{bai2025qwen3vl}  & 8B & Agent & 42.23 & 79.06 & 71.86 & 80.63 & 68.52 & 61.82 & 64.65 \\

\midrule
\textbf{Ours M2 (SFT)} & 4B & Agent & 60.95 & 82.00 & 78.95 & 93.72 & 75.08 & 66.87 & 71.55 \\
\textbf{Ours M3 (DPO)} & 4B & Agent & 62.60 & 82.60 & 79.32 & 94.76 & 75.08 & 65.00 & 66.72 \\
\textbf{Ours M4 (GSPO)} & 4B & Agent & 61.18 & 83.87 & 78.25 & 94.24 & 74.46 & 67.20 & 64.83 \\

\bottomrule
\end{tabular*}
\end{table}

%% file: tables/localization.tex
\begin{table}[!htbp]
\centering
\small
\setlength{\tabcolsep}{4pt}
\renewcommand{\arraystretch}{1.15}

\caption{Training improves evidence-region coverage and one-call accuracy. One-call acc. denotes final-answer accuracy conditional on exactly one tool call. Changes in parentheses are relative to M1.}
\label{tab:localization}

\begin{tabularx}{\textwidth}{
@{}
l
*{6}{>{\centering\arraybackslash}X}
@{}
}
\toprule
& \multicolumn{3}{c}{\textbf{ZoomBench (845)}}
& \multicolumn{3}{c}{\textbf{Aperture-580}} \\
\cmidrule(lr){2-4}\cmidrule(lr){5-7}

\textbf{Model}
& \textbf{Contain} $\uparrow$
& \textbf{Tight} $\uparrow$
& \shortstack{\textbf{One-call} $\uparrow$}
& \textbf{Contain} $\uparrow$
& \textbf{Tight} $\uparrow$
& \shortstack{\textbf{One-call} $\uparrow$} \\
\midrule

M1 (base)
& 0.191 & 0.159 & 0.530
& 0.474 & 0.103 & 0.596 \\

M2 (SFT)
& 0.347 {\scriptsize(+15.6pp)}
& 0.173 {\scriptsize(+1.4pp)}
& 0.653 {\scriptsize(+12.3pp)}
& 0.707 {\scriptsize(+23.3pp)}
& 0.220 {\scriptsize(+11.7pp)}
& 0.686 {\scriptsize(+9.0pp)} \\

M3 (DPO)
& 0.347 {\scriptsize(+15.6pp)}
& 0.175 {\scriptsize(+1.6pp)}
& 0.663 {\scriptsize(+13.3pp)}
& 0.716 {\scriptsize(+24.2pp)}
& 0.220 {\scriptsize(+11.7pp)}
& 0.700 {\scriptsize(+10.4pp)} \\

M4 (GSPO)
& 0.307 {\scriptsize(+11.6pp)}
& 0.140 {\scriptsize(-1.9pp)}
& 0.650 {\scriptsize(+12.0pp)}
& 0.667 {\scriptsize(+19.3pp)}
& 0.185 {\scriptsize(+8.2pp)}
& 0.672 {\scriptsize(+7.6pp)} \\

\bottomrule
\end{tabularx}
\end{table}

%% file: tables/call-decisions.tex
\begin{table}[!htbp]
\centering
\small
\setlength{\tabcolsep}{5pt}
\renewcommand{\arraystretch}{1.15}

\caption{SFT improves observation recall while reducing unnecessary or redundant calls. The DPO and GSPO branches differ in how they balance recall and call reduction. Changes are relative to M1.}
\label{tab:call-decisions}

\begin{tabularx}{\textwidth}{
@{}
l
*{4}{>{\centering\arraybackslash}X}
@{}
}
\toprule
& \multicolumn{2}{c}{\textbf{ZoomBench (845)}}
& \multicolumn{2}{c}{\textbf{Aperture-580}} \\
\cmidrule(lr){2-3}\cmidrule(lr){4-5}

\textbf{Model}
& \textbf{Over-Call} $\downarrow$
& \textbf{Necessary Recall} $\uparrow$
& \textbf{Over-Call} $\downarrow$
& \textbf{Necessary Recall} $\uparrow$ \\
\midrule

M1 (base)
& 0.273
& 0.557
& 0.753
& 0.832 \\

M2 (SFT)
& 0.111 {\scriptsize(-16.2pp)}
& 0.813 {\scriptsize(+25.6pp)}
& 0.284 {\scriptsize(-46.9pp)}
& 0.871 {\scriptsize(+3.9pp)} \\

M3 (DPO)
& 0.087 {\scriptsize(-18.6pp)}
& 0.814 {\scriptsize(+25.7pp)}
& 0.284 {\scriptsize(-46.9pp)}
& 0.884 {\scriptsize(+5.2pp)} \\

M4 (GSPO)
& 0.087 {\scriptsize(-18.6pp)}
& 0.731 {\scriptsize(+17.4pp)}
& 0.204 {\scriptsize(-54.9pp)}
& 0.831 {\scriptsize(-0.1pp)} \\

\bottomrule
\end{tabularx}
\end{table}

%% file: tables/controlled-recovery.tex
\begin{table}[!htbp]
\centering
\small
\setlength{\tabcolsep}{4pt}
\renewcommand{\arraystretch}{1.15}

\caption{Trained models retry more often and more frequently recover both localization and a correct answer from the same bad first crop. Relocalization requires a later crop with $\mathrm{cov}\geq 0.5$, and absolute recovery also requires a correct final answer. All rates use all 150 controlled examples per benchmark. Parenthesized changes are percentage points relative to M1.}
\label{tab:controlled-recovery}

\begin{tabularx}{\textwidth}{
@{}
l
*{6}{>{\centering\arraybackslash}X}
@{}
}
\toprule
& \multicolumn{3}{c}{\textbf{ZoomBench (845)}}
& \multicolumn{3}{c}{\textbf{Aperture-580}} \\
\cmidrule(lr){2-4}\cmidrule(lr){5-7}

\textbf{Model}
& \textbf{Retry} $\uparrow$
& \textbf{Relocalization} $\uparrow$
& \textbf{Abs. recovery} $\uparrow$
& \textbf{Retry} $\uparrow$
& \textbf{Relocalization} $\uparrow$
& \textbf{Abs. recovery} $\uparrow$ \\
\midrule

M1(base)
& 82.70\% & 52.70\% & 28.70\%
& 95.30\% & 70.00\% & 40.70\% \\

M2(SFT)
& 96.70\% {\scriptsize(+14.00pp)}
& 56.70\% {\scriptsize(+4.00pp)}
& 36.00\% {\scriptsize(+7.30pp)}
& 99.30\% {\scriptsize(+4.00pp)}
& 89.30\% {\scriptsize(+19.30pp)}
& 56.00\% {\scriptsize(+15.30pp)} \\

M3(DPO)
& 96.70\% {\scriptsize(+14.00pp)}
& 58.70\% {\scriptsize(+6.00pp)}
& 35.30\% {\scriptsize(+6.60pp)}
& 99.30\% {\scriptsize(+4.00pp)}
& 90.00\% {\scriptsize(+20.00pp)}
& 57.30\% {\scriptsize(+16.60pp)} \\

M4(GSPO)
& 96.00\% {\scriptsize(+13.30pp)}
& 58.00\% {\scriptsize(+5.30pp)}
& 36.00\% {\scriptsize(+7.30pp)}
& 99.30\% {\scriptsize(+4.00pp)}
& 92.70\% {\scriptsize(+22.70pp)}
& 57.30\% {\scriptsize(+16.60pp)} \\

\bottomrule
\end{tabularx}
\end{table}

%% file: tables/abstention.tex
\begin{table}[!htbp]
\centering
\small
\setlength{\tabcolsep}{2.5pt}
\renewcommand{\arraystretch}{1.15}

\caption{SFT improves abstention on unanswerable examples, and DPO retains this gain. The GSPO branch abstains less often and fabricates more answers.}
\label{tab:abstention}

\begin{tabular*}{\textwidth}{
@{\extracolsep{\fill}}
lrrrrr
@{}
}
\toprule
& \multicolumn{3}{c}{\textbf{Outcome Counts}}
& \multicolumn{2}{c}{\textbf{Rates}} \\
\cmidrule(lr){2-4}\cmidrule(lr){5-6}

\textbf{Model}
& \shortstack{\textbf{Appropriate}\\\textbf{abst.}}
& \textbf{Fabric.}
& \textbf{Repet.}
& \shortstack{\textbf{Abst.}\\\textbf{rate} $\uparrow$}
& \shortstack{\textbf{Halluc.}\\\textbf{rate} $\downarrow$} \\
\midrule

M1 (base) & 52 & 21 & 1 & 70.27\% & 28.38\% \\
M2 (SFT) & 59 & 15 & 0 & 79.73\% & 20.27\% \\
M3 (DPO) & 59 & 15 & 0 & 79.73\% & 20.27\% \\
M4 (GSPO) & 44 & 30 & 0 & 59.45\% & 40.55\% \\

\bottomrule
\end{tabular*}
\end{table}

%% file: sections/statements.tex
\listedsection{LLM Usage Statement}

We used ChatGPT (OpenAI), Claude (Anthropic) and Qwen3.8-max (Qwen) for manuscript formatting,
language polishing, code implementation and debugging, data organization and
visualization, and supplementary implementation and hyperparameter descriptions.
LLM teacher models also generated training data and trajectories, as described in
the methodology. The authors reviewed and verified all research decisions,
experimental results, technical claims, and final manuscript content.

\listedsection{Ethics Statement}
A zoomed crop can reveal identifying details that are difficult to see in the full image. Anyone sharing a crop should consider what it reveals about the people depicted. Super-resolution can make a character appear legible even when the source pixels do not support that reading. The agent prompt directs the model to check such details against the raw crop and to abstain when the available evidence is insufficient.

\listedsection{Reproducibility Statement}
Appendix~\ref{app:impl} describes the zoom tool, the three-call budget, the crop and magnification rules, and the settings used to evaluate the model checkpoints. Appendix~\ref{app:prompts} gives the agent prompt and action formats. For the controlled recovery study, each model receives the same faulty first crop for a given question. Appendix~\ref{app:recovery} specifies the sample sizes and the criteria for recovering the evidence and answering correctly.

\listedsection{Research Scope, Data Use, and Institutional Disclaimer}
ZoomLynx studies how a vision-language model selects and uses local visual evidence under a fixed observation budget. The quantitative experiments evaluate image question answering. Evidence about cross-image reasoning is limited to the qualitative examples in Appendix~\ref{app:qualitative}. Third-party images, datasets, and model weights remain subject to their respective licenses and terms of use.

The views, methods, findings, and conclusions presented in this paper are those of the authors and do not represent the official positions, product plans, or commercial commitments of XPeng Inc. or the other affiliated institutions. Institutional affiliations do not imply endorsement of downstream applications.

%% file: sections/appendix.tex
\clearpage
\beginappendix

\section{LLM Usage Statement}
\label{app:llm}

We used ChatGPT (OpenAI), Claude (Anthropic), and Qwen3.8-Max for manuscript editing, formatting, implementation and debugging, organizing experimental results, and preparing figures and hyperparameter descriptions. Qwen3.8-Max also generated candidate questions and repaired observation trajectories. GPT-5.5 checked candidate benchmark items and judged open-ended answers and ambiguous observations. The authors reviewed the manuscript, results, and technical claims.
\section{Complete Experimental Results}
\label{app:results}
\subsection{Benchmarks, Metrics, and Evaluation Protocol}
\label{app:ap-eval}

\paragraph{Conditions.} In the Direct condition the model sees the image and the
question with no tool and produces a single answer. In the Agent condition it
sees the zoom protocol prompt and the tool schema, may call the tool up to three
times, and terminates with the JSON contract of Appendix~\ref{app:tool}. Both
conditions use greedy decoding at temperature 0.

\paragraph{Geometric metrics.} Let $c$ denote the crop and $r$ the annotated evidence region. Coverage is $\mathrm{area}(c\cap r)/\mathrm{area}(r)$. A crop contains the evidence if its coverage is one. A containing crop is tight if $\mathrm{area}(c)/\mathrm{area}(r)\leq 2$. A crop that shows only part of the evidence therefore fails both tests. One-call accuracy measures final-answer accuracy on the subset where a model made exactly one call. This subset can differ between models.

\paragraph{Decision metrics.} On Aperture-580 the over-call rate is the fraction
of directly answerable items that still receive a call, and necessary recall is
the fraction of necessary items that receive at least one call. On ZoomBench,
which provides no direct or necessary labels but does provide an official crop
region for every item, all items are treated as necessary and necessary recall
reduces to the call rate; the over-call rate there is the fraction of items with
at least two calls whose last call repeats an earlier region or adds no new
image or super-resolution detail, with GPT-5.5 adjudicating uncertain cases. The
two benchmarks use different populations, so over-call rates are compared only
within a benchmark.

\paragraph{Recovery protocol.} We sample 150 Necessary items from Aperture-580 and 150 items from ZoomBench. Each model receives the same first crop for a given item, with evidence coverage at most 0.2. The model chooses all subsequent calls. A retry is any later call; relocalization requires a later crop with coverage at least 0.5. Absolute recovery requires both relocalization and a correct answer. The coverage-improvement measure instead requires a gain of at least 0.3 over the first crop and a correct answer. All rates use the full 150-item sample for each benchmark.

\paragraph{Abstention audit.} On the 74 Stop items, we distinguish a valid abstention with an appropriate reason, an unsupported answer, and malformed or repeated output. We separately check whether the model abstains on answerable items.

\paragraph{Answer judging.} For open-ended Chinese answers on Aperture-580, GPT-5.5 assigns one of three judgments: equivalent, partial, or different. Only an equivalent answer receives credit. The answer matcher treats an empty answer or refusal as different; the Stop audit evaluates abstention separately. Public benchmarks use option matching or normalized answer matching. Scorer choice matters: scores on the Stop subset can differ by about 22 percentage points.

\subsection{Per-Bucket Accuracy on Aperture-580}
\label{app:per-bucket}

All three trained models improve on M1 in the Direct, Necessary, and Multi-region buckets (Table~\ref{tab:per-bucket}). M3 achieves the highest Multi-region accuracy (58.4\% versus 40.6\% for M1). Stop shows a different pattern: M2 and M3 both reach 79.7\%, while M4 falls to 59.5\%. Conditional accuracy (CondAcc) uses only items on which the model made the expected observation decision; models can therefore have different denominators.

\begin{table}[!htbp]
\centering
\small
\caption{Answer accuracy (\%) by Aperture-580 bucket. The buckets contain 162 Direct, 243 Necessary, 101 Multi-region, and 74 Stop items. CondAcc denotes accuracy conditional on the expected observation decision.}
\label{tab:per-bucket}
\setlength{\tabcolsep}{2.4pt}
\begin{tabularx}{\linewidth}{@{}l*{5}{>{\centering\arraybackslash}X}@{}}
\toprule
\textbf{Model} & \textbf{Direct} & \textbf{Necess.} & \textbf{Multi} & \textbf{Stop} & \textbf{CondAcc} \\
\midrule
M1 & 66.7 & 63.4 & 40.6 & 70.3 & 57.0 \\
M2 & 77.2 & 67.5 & 56.4 & 79.7 & 69.1 \\
M3 & 78.4 & 69.5 & 58.4 & 79.7 & 70.8 \\
M4 & 76.5 & 68.3 & 52.5 & 59.5 & 69.3 \\
\bottomrule
\end{tabularx}
\end{table}

\subsection{Controlled Recovery under a Shared Faulty First Crop}
\label{app:recovery}

A later crop can reveal useful evidence even if its coverage remains below 0.5. Table~\ref{tab:recovery-ci} uses improvement over the initial crop as the localization criterion. It thus captures some partial recoveries that the absolute threshold excludes.

\begin{table}[!htbp]
\centering
\footnotesize
\caption{Coverage-improvement recovery under the shared faulty first crop. The
criterion requires a later crop whose coverage exceeds that of the bad crop by at
least 0.3, together with a correct final answer. Rates are computed over the full controlled evaluation set.}
\label{tab:recovery-ci}
\setlength{\tabcolsep}{2.5pt}
\begin{tabularx}{\linewidth}{@{}l*{4}{>{\centering\arraybackslash}X}@{}}
\toprule
\textbf{Model} & \multicolumn{2}{c}{\textbf{ZoomBench}} & \multicolumn{2}{c}{\textbf{Aperture-580}} \\
\cmidrule(lr){2-3}\cmidrule(lr){4-5}
 & \textbf{Rate} & \textbf{$\Delta$ vs M1} & \textbf{Rate} & \textbf{$\Delta$ vs M1} \\
\midrule
M1 & 32.0 & --- & 41.3 & --- \\
M2 & 42.7 & +10.7 & 58.0 & +16.7 [6.7, 25.3] \\
M3 & 42.7 & +10.7 & 60.0 & +18.7 [9.3, 28.0] \\
M4 & 42.7 & +10.7 & 57.3 & +16.0 [6.0, 25.3] \\
\bottomrule
\end{tabularx}
\end{table}

On ZoomBench, all three trained models improve on M1 by 10.7 percentage points under this criterion (paired 95\% bootstrap CI [2.0, 19.3]). On Aperture-580 the gains range from 16.0 to 18.7 points; all three intervals exclude zero.

\paragraph{Additional recovery measurements.}
Table~\ref{tab:recovery-extended} compares accuracy on the same questions with and without the injected crop. After injection, relocalization and answering correctly are separate outcomes. On Aperture-580, for example, M4 has the highest relocalization rate (92.7\%), whereas M3 has the highest answer accuracy (63.3\%).

\begin{table}[!htbp]
\centering
\small
\caption{Accuracy and localization on the paired recovery sets (\%, 150 items per benchmark). Original accuracy uses the ordinary rollout. The remaining columns use the shared faulty first crop. Relocalization requires coverage of at least 0.5; the final column uses the coverage-improvement criterion in Table~\ref{tab:recovery-ci}.}
\label{tab:recovery-extended}
\setlength{\tabcolsep}{2.5pt}
\begin{tabularx}{\linewidth}{@{}l*{4}{>{\centering\arraybackslash}X}@{}}
\toprule
\textbf{Model} & \textbf{Original Acc.} & \textbf{Acc. after bad crop} &
\textbf{Relocali\-zation} & \textbf{Recovery (gain)} \\
\midrule
\multicolumn{5}{@{}l}{\emph{Aperture-580}} \\
M1 & 57.3 & 48.0 & 70.0 & 41.3 \\
M2 & 58.0 & 60.7 & 89.3 & 58.0 \\
M3 & 58.7 & 63.3 & 90.0 & 60.0 \\
M4 & 62.7 & 60.7 & 92.7 & 57.3 \\
\midrule
\multicolumn{5}{@{}l}{\emph{ZoomBench}} \\
M1 & 48.7 & 49.3 & 52.7 & 32.0 \\
M2 & 62.7 & 63.3 & 56.7 & 42.7 \\
M3 & 66.7 & 62.0 & 58.7 & 42.7 \\
M4 & 66.0 & 62.0 & 58.0 & 42.7 \\
\bottomrule
\end{tabularx}
\end{table}

\subsection{Stop Behavior and Abstention Diagnostics}
\label{app:stop-diagnostics}

A structured stop is appropriate only when the image cannot support a definite answer. Table~\ref{tab:stop-diagnostics} separates abstention from unsupported answers and terminal-format errors.

\begin{table}[!htbp]
\centering
\small
\caption{Stop and abstention diagnostics on the frozen Stop subset ($n=74$).
Entries are counts with rates in parentheses. Valid abstention is a structured
stop accepted by the judge; unsupported answer is a non-empty answer without
supporting evidence; malformed/repeated is a terminal-format violation or a
repeated final-answer output.}
\label{tab:stop-diagnostics}
\begin{tabularx}{\linewidth}{@{}l*{3}{>{\centering\arraybackslash}X}@{}}
\toprule
\textbf{Model} & \textbf{Valid abstention} & \textbf{Unsupported answer} &
\textbf{Malformed / repeated} \\
\midrule
M1 & 52 (70.3) & 21 (28.4) & 1 (1.4) \\
M2 & 59 (79.7) & 15 (20.3) & 0 (0.0) \\
M3 & 59 (79.7) & 15 (20.3) & 0 (0.0) \\
M4 & 44 (59.5) & 30 (40.5) & 0 (0.0) \\
\bottomrule
\end{tabularx}
\end{table}

M2 and M3 each abstain appropriately on 59 of 74 items, seven more than M1. M4 gives 30 unsupported answers, twice as many as M2 or M3. The only repeated or malformed response comes from M1 (1/74, 1.4\%).

\subsection{Contribution of the Enhanced View}
\label{app:sr-ablation}

We compare three observations for the same questions: a raw crop plus a Real-ESRGAN view, a raw crop plus a Lanczos-resampled view, and the raw crop alone. The scale rule in Table~\ref{tab:scale} determines which crops receive an enlarged view. Relative to Real-ESRGAN, Lanczos changes accuracy by $-0.47$ points (95\% CI $[-2.25,+1.30]$) across all items and by $+0.79$ points ($[-3.57,+5.16]$) on items enlarged in at least one condition. Removing the enlarged view changes accuracy by $-1.18$ points overall ($[-2.96,+0.59]$) and $-4.37$ on the latter subset ($[-8.73,0.00]$). The largest estimated loss occurs where the tool enlarged the crop. The intervals allow a zero change, and the experiment does not establish an advantage for one upsampling method. At 80\% power the minimum detectable effects are 2.5 and 6.2 points for the two populations.

\subsection{Aiming versus Tightness}
\label{app:aiming}

On 193 Necessary items where both M1 and M2 call the tool, we compare their first crops on the same questions. The crop center falls inside the evidence region more often for M2 (0.860 versus 0.591). Yet mean IoU falls from 0.419 to 0.356, and the fraction with IoU at least 0.5 falls from 0.409 to 0.301. The share of crops more than twice the evidence area rises from 56.0\% to 60.6\%, while the share smaller than the evidence region falls from 13.0\% to 9.8\%. The median crop-to-evidence area ratios remain similar (2.56 and 2.49). Thus a better-placed center can coexist with a lower IoU when the crop includes more context.

\subsection{Prompt Robustness}
\label{app:robustness}

Training matches the prompt language to the question. For English ZoomBench questions, the main paper uses the English canonical prompt T3 (47.10\% for M1, 60.95\% for M2). Switching to the Chinese canonical prompt P0 raises both scores by about two points while leaving the M2--M1 gap similar (Table~\ref{tab:prompt-robust}).

Across P0--P4, accuracy spans 27.2--55.5\% for M1 and 61.2--63.1\% for M2. Relative to P0, the M1 differences on P1--P4 are $+6.0$, $-8.9$, $-22.3$, and $+0.5$ points; the paired 95\% intervals for P1 and P4 are $[+3.1,+9.0]$ and $[-2.6,+3.6]$. For M2 the corresponding differences are $-0.1$, $-0.8$, $-1.9$, and $-1.4$ points, with all four intervals including zero. M2 exceeds M1 on all four held-out wordings, with a mean gain of 18.7 points (95\% CI [16.1, 21.3]). The gain persists across the tested variants; the study does not establish equivalence across prompts.

\begin{table}[!htbp]
\centering
\footnotesize
\setlength{\tabcolsep}{4pt}
\renewcommand{\arraystretch}{1.05}
\caption{ZoomBench prompt variants. T3 and P0 are the training prompts for English and Chinese questions, respectively. P1--P4 were held out from training, model selection, and tuning. The last column gives the paired accuracy gain over M1 in percentage points. Across the held-out variants, every gain is positive, the mean and lower 95\% confidence bound exceed zero, and the mean exceeds $0.7$ times the P0 gain.}
\label{tab:prompt-robust}
\begin{tabularx}{\linewidth}{@{}l X r r r@{}}
\toprule
\textbf{Prompt} & \textbf{Variant} & \textbf{Base} & \textbf{M2} & \textbf{$\Delta$} \\
\midrule
T3 & English canonical (in-distribution) & 47.1 & 61.0 & $+13.9$ \\
P0 & Chinese canonical, English questions & 49.5 & 63.1 & $+13.6$ \\
P1 & Chinese concise & 55.5 & 63.0 & $+7.5$  \\
P2 & Chinese reordered & 40.6 & 62.3 & $+21.7$ \\
P3 & English equivalent (short) & 27.2 & 61.2 & $+34.0$ \\
P4 & English minimal & 49.9 & 61.7 & $+11.7$ \\
\midrule
\multicolumn{5}{@{}l}{\textbf{Mean gain (P1--P4)}: $+18.7$ (95\% CI $[16.1,21.3]$)} \\
\bottomrule
\end{tabularx}
\end{table}

\subsection{Effects of Post-SFT Optimization}
\label{app:posttrain}

M3 (DPO) and M4 (GSPO) start from the same supervised checkpoint, M2. Their training data and objectives both differ. Table~\ref{tab:posttrain} reports the net change for each branch across seven benchmarks.

\begin{table}[!htbp]
\centering
\small
\caption{Change relative to M2, in percentage points. Negative values mean the
post-SFT branch is worse than the SFT model on that benchmark. Column abbreviations follow Table~\ref{tab:benchmark-results}.}
\label{tab:posttrain}
\setlength{\tabcolsep}{1.5pt}
\begin{tabularx}{\linewidth}{@{}l*{7}{>{\centering\arraybackslash}X}@{}}
\toprule
\textbf{Branch} & \textbf{ZB} & \textbf{H4K} & \textbf{H8K} & \textbf{VB} & \textbf{AV} & \textbf{MME} & \textbf{AP} \\
\midrule
M3 (DPO) & +1.65 & +0.60 & +0.37 & +1.04 & 0.00 & $-1.87$ & $-4.83$ \\
M4 (GSPO) & +0.23 & +1.87 & $-0.70$ & +0.52 & $-0.62$ & +0.33 & $-6.72$ \\
\bottomrule
\end{tabularx}
\end{table}

DPO improves four benchmark scores, leaves ActiView unchanged, and lowers MME-RealWorld-Lite and Aperture-580. On Aperture-580, necessary recall increases from 0.871 to 0.884 while the direct over-call rate (0.284), appropriate abstention (79.73\%), and crop tightness (0.220) stay the same. M2 already contains the evidence in 0.707 of its first crops but produces a tight crop on only 0.220, leaving room to improve localization precision.

GSPO reduces the direct over-call rate to 0.204, alongside lower necessary recall (0.831), appropriate abstention (59.45\%), and crop tightness (0.185). The call-cost term favors fewer observations, but these comparisons do not isolate its effect. GSPO samples eight responses per prompt and requires reward variation within a group for a nonzero update. In this run, 64.5\% of groups varied in reward, above the configured 35\% minimum; the other 35.5\% contributed no within-group advantage (about five to six of 16 groups per batch). Both post-SFT branches used fewer examples and optimization steps than SFT, limiting conclusions about longer training.

\subsection{Over-Call and Repeated-Observation Analysis}
\label{app:overcall-taxonomy}

For trajectories with multiple tool calls, we classify the final call by its
geometric relation to earlier crops. A call is a \emph{repeat} when its
intersection-over-union with a previous crop is at least 0.8. It is a
\emph{refinement} when it is fully contained in an earlier crop and its area is
less than 0.7 times that crop's area. It is a \emph{new-area observation} when
its IoU with every previous crop is below 0.3. Overlaps with IoU between 0.3 and
0.8 that are not refinements remain \emph{ambiguous}: geometric overlap alone
cannot determine whether the new crop contributes information.

We resolve ambiguous calls with GPT-5.5, providing the question, the preceding
crop, and the final crop, and asking whether the final crop adds visual evidence
not already visible. Of the 26 ambiguous final calls on Aperture-580, 20 were
judged uninformative and 6 informative. A repeated call, or an ambiguous call
judged uninformative, counts as an unproductive final observation; refinements
and new-area observations do not. The unproductive rate is 9.2\% (13/141) for M1,
2.1\% (3/141) for M2, 3.4\% (5/147) for M3, and 6.6\% (8/121) for M4, where the
denominator is the number of trajectories with at least two tool calls.

\begin{table}[!htbp]
\centering
\footnotesize
\setlength{\tabcolsep}{3pt}
\caption{Final-call classification on Aperture-580 for trajectories with at
least two tool calls. Ambiguous calls are adjudicated by GPT-5.5; ``uninf.''
counts those judged to add no visual evidence. Unproductive = repeat + ambiguous
uninformative.}
\label{tab:overcall}
\begin{tabularx}{\linewidth}{@{}X r r r r@{}}
\toprule
 & \textbf{M1} & \textbf{M2} & \textbf{M3} & \textbf{M4} \\
\midrule
Multi-call      & 141 & 141 & 147 & 121 \\
Repeat          &   7 &   1 &   0 &   1 \\
Refinement      &  14 &  76 &  83 &  68 \\
New-area        & 112 &  61 &  58 &  43 \\
Ambiguous       &   8 &   3 &   6 &   9 \\
\quad uninf.    &   6 &   2 &   5 &   7 \\
Unproductive    &  13 &   3 &   5 &   8 \\
Rate            & 9.2\% & 2.1\% & 3.4\% & 6.6\% \\
\bottomrule
\end{tabularx}
\end{table}

\section{Agentic Training and Evaluation Settings}
\label{app:impl}

\subsection{Zoom Tool Protocol}
\label{app:tool}

The policy can inspect any currently visible image using one tool, \texttt{zoom\_and\_super\_resolve}. A call specifies the image index and the bounding box to inspect. After a crop is returned, the model can inspect it again or return to the original image.

\paragraph{Observations and budget.} Each episode allows three attempted calls. Invalid arguments, an invalid image index, or a region that is too small still consume a call. A rejected call returns no image. A successful call returns the lossless crop and, when its size warrants enhancement, a magnified view. If enhancement fails, the raw crop remains available. Each response reports the number of calls left. A crop taken from a previous observation is mapped back to original-image coordinates.

\paragraph{Reasoning and stopping.} Before inspecting a region, the model identifies the evidence it has, the detail it still needs, and where to look. It reasons in the language of the question and can use only images already observed. At termination it answers or gives a reason for abstaining, such as insufficient evidence, an absent target, an ambiguous query, a need for external knowledge, an exhausted budget, or an unrecoverable tool error. The answer scorer reads the final answer separately from the accompanying reasoning. Appendix~\ref{app:prompts} gives the prompt and action formats.

\subsection{Image Cropping and Super-Resolution}
\label{app:sr}

\paragraph{Raw crop.} We apply EXIF orientation, convert the image to RGB, and clamp the selected box to the image boundaries. We store the crop losslessly as PNG and reject a crop whose shorter side is under 8 pixels. For a nested crop, the parent image and its coordinates determine the corresponding region in the original image.

\paragraph{Super-resolution.} We use the official Real-ESRGAN $2\times$ and $4\times$ checkpoints in fp16 on CUDA. The RRDBNet backbone has 23 residual-in-residual dense blocks, 64 features, growth width 32, and residual scale 0.2. The $2\times$ model uses pixel unshuffle at its input. The service accepts and returns lossless PNG images with a 600-second client timeout.

\paragraph{Automatic scale selection.} The policy chooses a region; the executor chooses the magnification. Table~\ref{tab:scale} lists the thresholds in priority order. When neither enlargement fits the size limit, the raw crop is the sole observation.

\begin{table}[!htbp]
\centering
\small
\caption{Automatic magnification by crop size. The first applicable row determines the scale; each enlarged output is limited to 2048 pixels on its long side.}
\label{tab:scale}
\begin{tabularx}{\linewidth}{@{}lXl@{}}
\toprule
\textbf{Short side} & \textbf{Output guard} & \textbf{Scale} \\
\midrule
$<112$ & $4\times$ long side $\le 2048$ & $4\times$ \\
$<224$ & $2\times$ long side $\le 2048$ & $2\times$ \\
otherwise & --- & $1\times$ (no request) \\
\bottomrule
\end{tabularx}
\end{table}

\paragraph{Evidence attribution.} Real-ESRGAN uses fixed weights in evaluation mode without sampling. The raw crop is the source for factual details: the prompt cautions against reading characters that appear only in the enhanced view. Appendix~\ref{app:sr-ablation} compares raw-only and enlarged observations.

\subsection{Training Stages}
\label{app:train}

Supervised fine-tuning produces M2. Direct preference optimization (M3) and group sequence policy optimization (M4) independently start from M2. We merge the M3 preference adapter into the base weights before evaluation. We use the same super-resolution service for collecting training observations and evaluating the policy.

\subsection{Evaluation Settings}
\label{app:evalset}

We evaluate the ZoomLynx checkpoints with greedy decoding on the same backend and tool service (Table~\ref{tab:serve}).

\begin{table}[!htbp]
\centering
\small
\caption{Serving and decoding settings for the ZoomLynx evaluations.}
\label{tab:serve}
\begin{tabularx}{\linewidth}{@{}l>{\raggedright\arraybackslash}X@{}}
\toprule
\textbf{Setting} & \textbf{Value} \\
\midrule
Backend & vLLM, OpenAI-compatible server \\
Precision / parallelism & bf16, tensor parallelism 1 \\
Max model length & 49,152 tokens \\
Max concurrent sequences & 64 \\
GPU memory utilization & 0.85 \\
Images per prompt & at most 40 \\
Tool parsing & Hermes tool-call parser, auto tool choice \\
Temperature / top-$p$ & 0 / 1.0 \\
Max new tokens per turn & 2,048 \\
Tool budget & 3 calls \\
Super-resolution service & Real-ESRGAN, fp16, concurrency 2 \\
\bottomrule
\end{tabularx}
\end{table}

\paragraph{Statistics.} We use 10,000 paired bootstrap resamples. Each draw uses the same item indices across models and retains each model's trajectory for that item. Repeated serving sessions with an identical configuration differed by 0.3--1.8 accuracy points. The power estimates for the enhanced-view comparison are given in Appendix~\ref{app:sr-ablation}.

\section{Training Examples and Supervision}
\label{app:data}
\label{app:repair}
Table~\ref{tab:buckets} defines the target action for each evidence state. A crop that selects the wrong object, misses the target, or shows only part of the required evidence can call for a retry. A wide crop containing the evidence, a malformed call, and a copied annotation box are excluded from this definition of a localization failure.

\begin{table}[!htbp]
\centering
\small
\caption{\textbf{Supervision targets.} The observed evidence determines the action to supervise.}
\label{tab:buckets}
\begin{tabularx}{\textwidth}{@{}lXX@{}}
\toprule
\textbf{Bucket} & \textbf{Evidence state} & \textbf{Supervised behavior} \\
\midrule
Direct & The full image already answers the question. & Answer without calling the tool. \\
Necessary-Zoom & The answer requires a local detail, and the first crop is well localized. & Call Zoom once on the evidence region, then answer. \\
Recovery & The first crop is uninformative or misses the target. & Retry from the original image, relocalize, and answer from the new observation. \\
Multi-region & The evidence is spread over separated regions. & Issue several crops over distinct regions and combine the local readings. \\
Stop/Abstain & No available observation supports a reliable answer. & Emit a structured stop with a reason instead of guessing. \\
\bottomrule
\end{tabularx}
\end{table}

\begin{figure}[!htbp]
  \centering
  \includegraphics[width=1.0\linewidth]{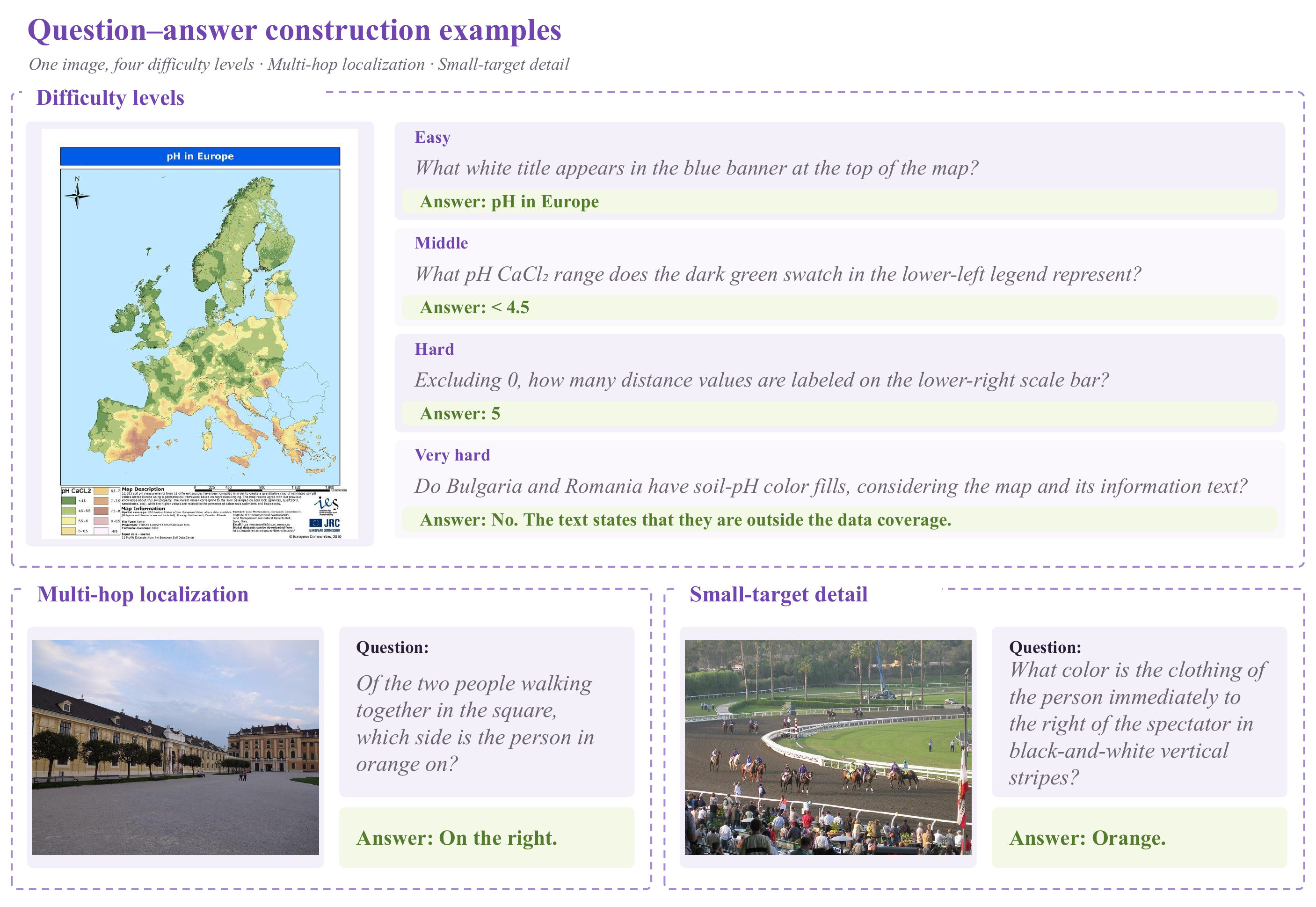}
  \caption{\textbf{Question--answer examples.} Four questions probe the same map at different scales. In the lower panels, a visual clue identifies a walker or singles out one spectator among several.}
  \label{fig:qa-construction}
  \par\medskip
  \nextfloat
  \includegraphics[width=1.0\linewidth]{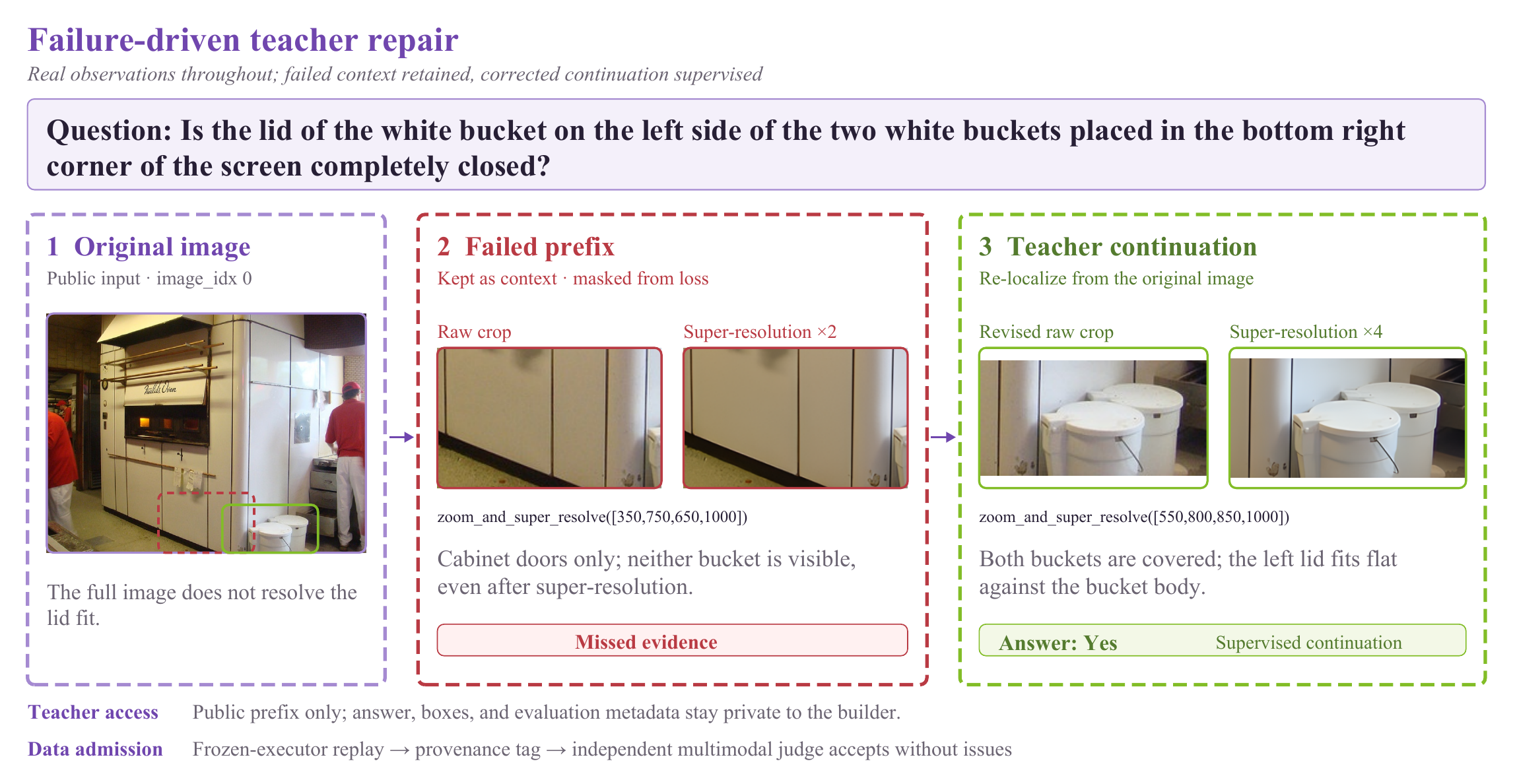}
  \caption{\textbf{Repair after a missed crop.} Enlarging the cabinet doors cannot resolve the question. The corrected crop shows both buckets and makes their lid fit visible.}
  \label{fig:repair-trajectory-supp}
\end{figure}

\begin{appendixcase}
\phantomsection{\sffamily\bfseries\color{templatetitle}Worked examples.}\enspace
\label{app:traj}
In Figure~\ref{fig:qa-construction}, four questions ask about the soil-pH map's title, legend, scale, and data coverage. Two more identify a walker in orange and a spectator next to the person in stripes. In Figure~\ref{fig:repair-trajectory-supp}, the teacher returns to the original image after a crop shows only cabinet doors. Supervision resumes at the corrected crop, where both lids are visible.
\end{appendixcase}

\FloatBarrier

\section{Qualitative Analysis}
\label{app:qualitative}
\subsection{Comparative Recovery Cases}
\label{app:comparative-cases}

In Figure~\ref{fig:recovery-cases}, M1 and M4 receive the same question and faulty first crop (coverage at most 0.2), then choose their next observations independently. Appendix~\ref{app:recovery} reports aggregate recovery rates.

\par\noindent\begin{minipage}{\linewidth}
  \centering
  \captionsetup{type=figure}
  \includegraphics[width=1.0\linewidth]{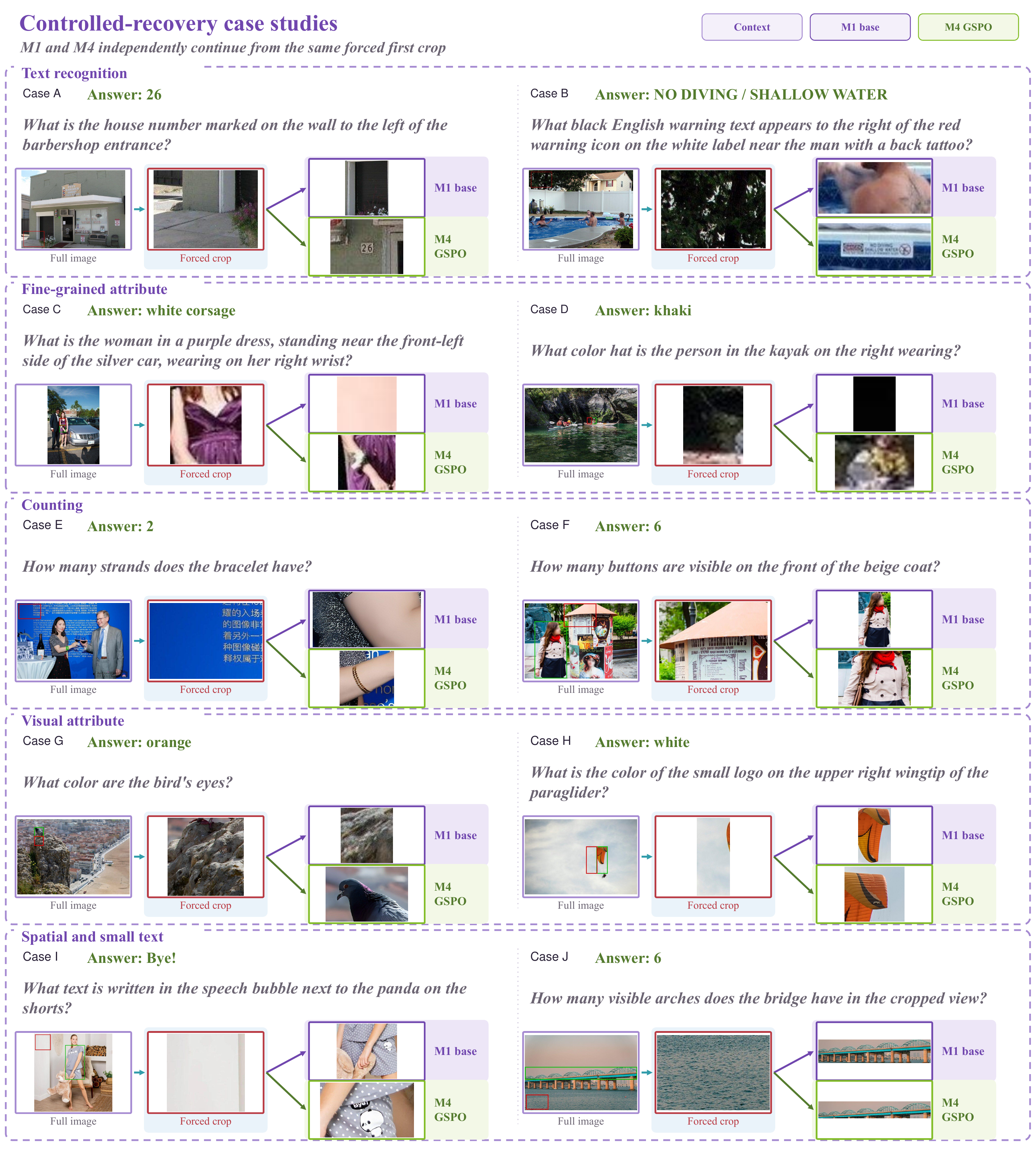}
  \caption{\textbf{Recovery from a shared forced crop.}
  Each row pairs two cases. M1 (base) often misses the evidence, while M4 (GSPO) relocates it and answers correctly.}
  \label{fig:recovery-cases}
\end{minipage}\par

\begin{tcbraster}[raster columns=2,raster equal height=rows,raster column skip=8pt,raster row skip=5pt,raster before skip=5pt,raster after skip=5pt]
\begin{appendixcase}
{\sffamily\bfseries\color{templatetitle}Case A: Door-number OCR.}\par\nobreak
The injected crop covers the side of the entrance and does not contain the door
number.  M1 remains close to this region and returns an incorrect reading.  M4
instead moves back to the entrance wall, centers the plaque, and reads
\textbf{26}.
\end{appendixcase}
\begin{appendixcase}
{\sffamily\bfseries\color{templatetitle}Case B: Warning-sign OCR.}\par\nobreak
The shared crop initially shows only an uninformative part of the pool area.
M1 continues to inspect irrelevant content, while M4 relocates the warning
label and reads \textbf{NO DIVING / SHALLOW WATER}.
\end{appendixcase}
\begin{appendixcase}
{\sffamily\bfseries\color{templatetitle}Case C: Wrist accessory.}\par\nobreak
The forced crop is blurred and does not establish the requested attribute.
M1 does not recover a useful observation.  M4 explicitly searches the person's
wrist, producing a tight crop that reveals a \textbf{white corsage}.
\end{appendixcase}
\begin{appendixcase}
{\sffamily\bfseries\color{templatetitle}Case D: Hat color.}\par\nobreak
The initial observation is uninformative for the hat.  M1 stays in the wrong
region, whereas M4 moves to the right-hand kayak and identifies the hat as
\textbf{khaki}.
\end{appendixcase}
\begin{appendixcase}
{\sffamily\bfseries\color{templatetitle}Case E: Strand counting.}\par\nobreak
The injected crop contains unrelated texture.  M1 remains anchored there and
cannot support the count.  M4 searches for the wrist, isolates the accessory,
and counts \textbf{two strands}.
\end{appendixcase}
\begin{appendixcase}
{\sffamily\bfseries\color{templatetitle}Case F: Button counting.}\par\nobreak
The shared crop covers background content rather than the coat.  M1 continues
to inspect the wrong region.  M4 relocates the coat front, frames the full
button column, and counts \textbf{six buttons}.
\end{appendixcase}
\begin{appendixcase}
{\sffamily\bfseries\color{templatetitle}Case G: Bird-eye color.}\par\nobreak
The first crop shows only rocks and plants.  M1 remains near this distractor,
while M4 moves to the bird's head and identifies the eye color as
\textbf{orange}.
\end{appendixcase}
\begin{appendixcase}
{\sffamily\bfseries\color{templatetitle}Case H: Paraglider-logo color.}\par\nobreak
The initial crop misses the relevant wingtip logo.  M1 fails to localize the
target, whereas M4 moves to the wingtip and determines that the logo is
\textbf{white}.
\end{appendixcase}
\begin{appendixcase}
{\sffamily\bfseries\color{templatetitle}Case I: Speech-bubble OCR.}\par\nobreak
The shared crop contains unrelated image content.  M1 remains trapped in that
region.  M4 searches for the graphic element, centers the speech bubble, and
reads \textbf{Bye!}.
\end{appendixcase}
\begin{appendixcase}
{\sffamily\bfseries\color{templatetitle}Case J: Bridge-arch counting.}\par\nobreak
The forced crop shows only part of the scene and cannot support a global count.
M1 does not recover the full bridge span.  M4 progressively reframes the
bridge and counts \textbf{six arches}.
\end{appendixcase}
\end{tcbraster}

\subsection{Multi-Image Observation and Cross-Image Evidence Integration}
\label{app:multi-view}

\begin{figure}[!htbp]
  \centering
  \includegraphics[width=1.0\linewidth]{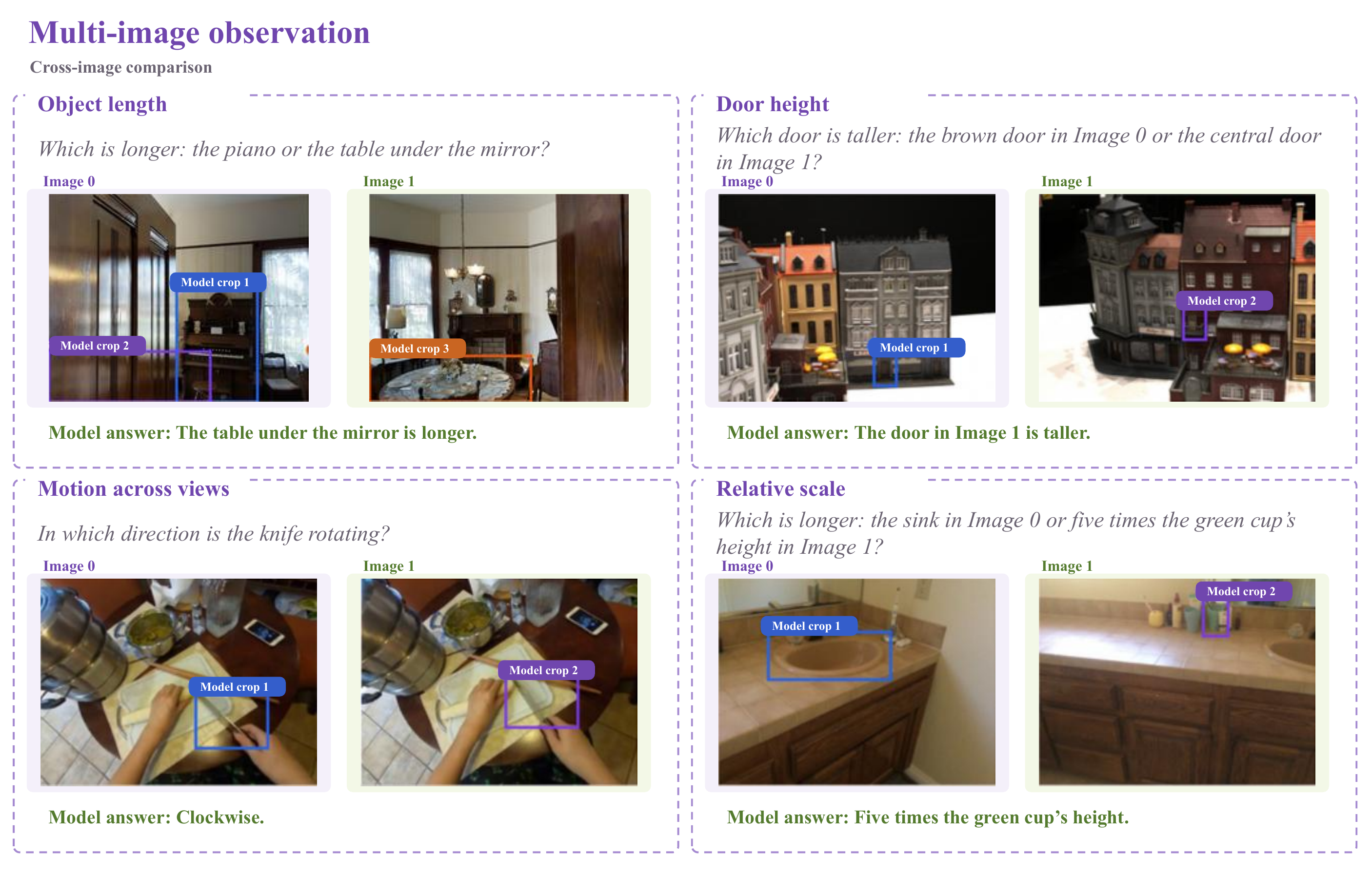}
\caption{Four successful cross-image comparisons. Numbered boxes indicate the order in which the policy inspected regions. Each example shows the source images, question, and final answer.}
  \label{fig:multi-view}
\end{figure}

\begin{appendixcase}
Figure~\ref{fig:multi-view} shows four successful episodes using a policy trained on single images. It selects a view at each step and integrates local evidence across views to compare lengths, door heights, the rotation of a knife, and the scale of a sink relative to a cup. The numbered boxes locate the selected regions in the input images.
\end{appendixcase}

These cases illustrate a capability of the observation interface; they do not measure how often the policy succeeds on multi-image questions.

\subsection{Multi-Region Observation}
\label{app:multi-region}

\begin{figure}[!htbp]
\centering
\includegraphics[width=1.0\linewidth]{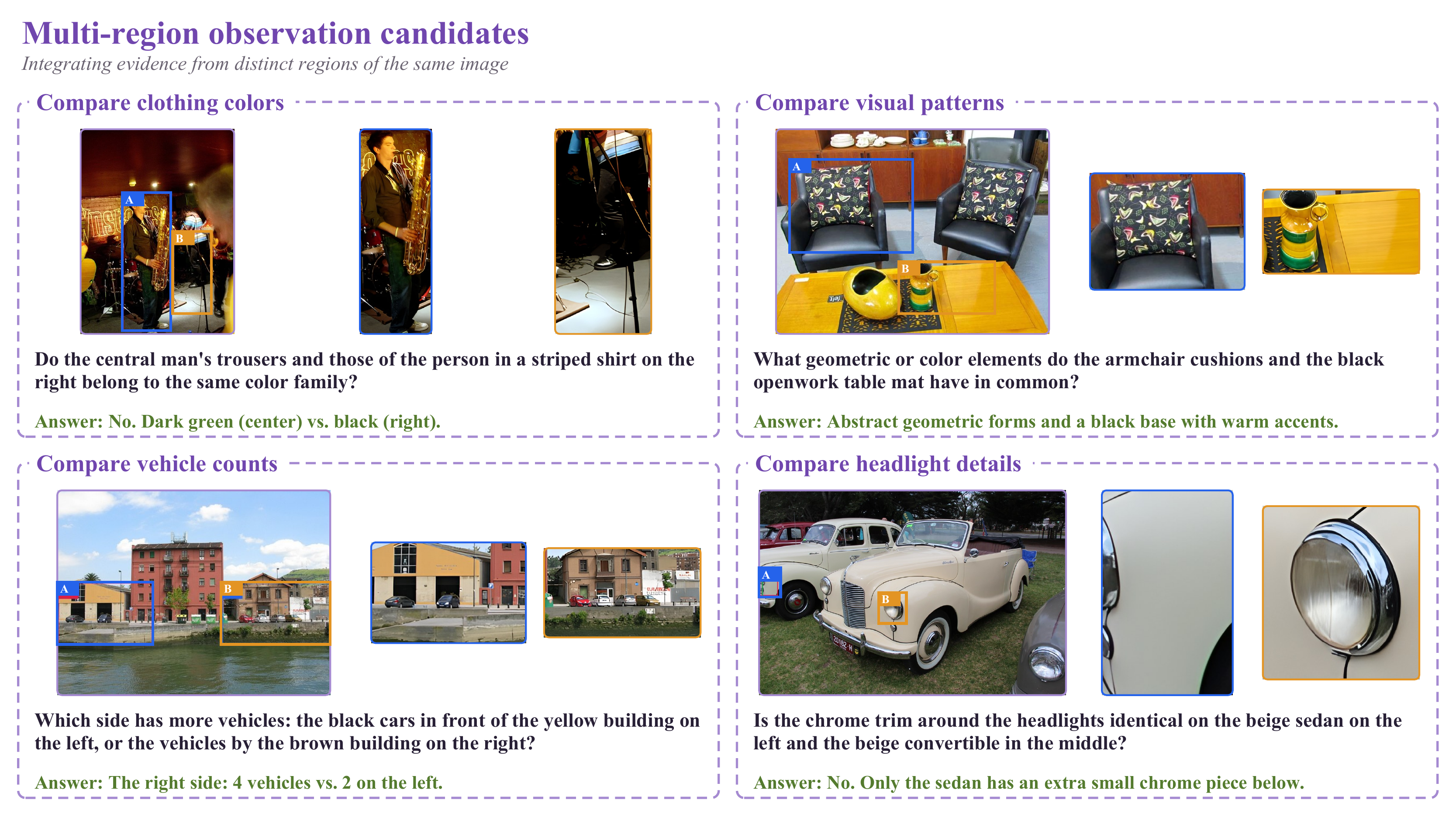}
\caption{Examples requiring evidence from two regions. Blue and orange outlines identify the two source regions and the corresponding crops. The questions ask about clothing colors, shared visual patterns, vehicle counts, and headlight details.}
\label{fig:multi-region}
\end{figure}

\begin{appendixcase}
The four training examples in Figure~\ref{fig:multi-region} require evidence from two locations in one image. Three trajectories include a Qwen teacher repair; the cushion-and-table-mat example was accepted as generated.
\end{appendixcase}

The selected regions and crops document how the training targets combine evidence across locations.

\FloatBarrier

\subsection{Recovery Failure and Stopping Behavior}
\label{app:recovery-failure-stop}

\begin{figure}[!htbp]
\centering
\includegraphics[width=1.0\linewidth]{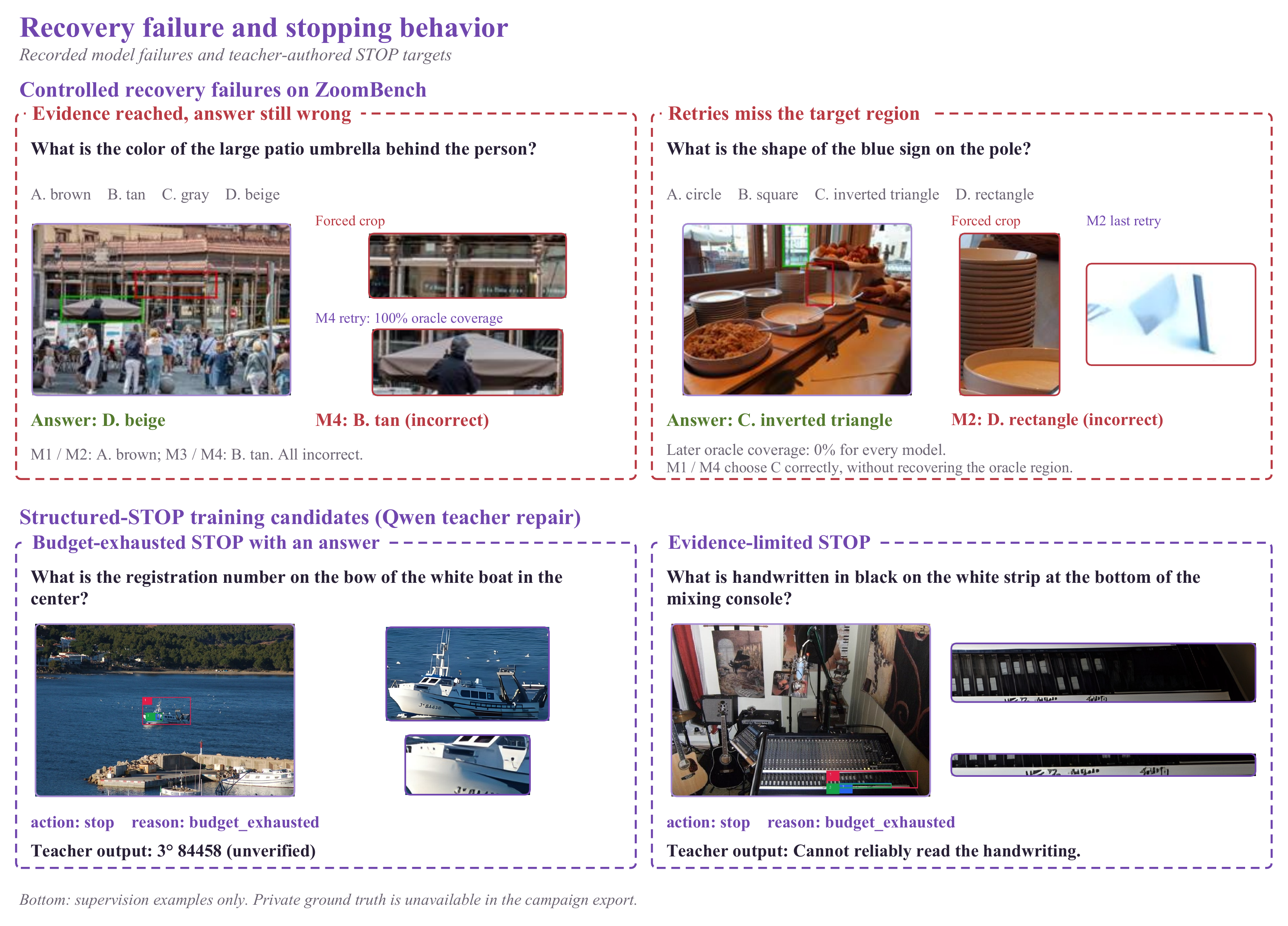}
\caption{Recovery failures in answer readout and localization (top), and teacher-authored structured-STOP targets (bottom).}
\label{fig:recovery-failure-stop}
\end{figure}

\begin{appendixcase}
Figure~\ref{fig:recovery-failure-stop} illustrates two reasons a recovery can fail. In the umbrella example, all four models later cover the evidence region but misidentify the color. In the sign example, no model relocates the evidence, although M1 and M4 select the correct option.
\end{appendixcase}

The answer alone cannot distinguish a successful recovery from a correct guess.

\begin{appendixcase}
The lower row compares two teacher-authored training targets. Both stop when the budget is exhausted. One still supplies a boat registration number, while the other abstains from reading an illegible label on a mixing console.
\end{appendixcase}

A stop action alone therefore does not establish abstention. No independent ground truth is available for these two targets.

\subsection{Failure Cases and Limitations}
\label{app:failures}

Table~\ref{tab:failures} links each failure to an observable symptom and a diagnostic. Overlapping the target, interpreting the crop correctly, and deciding whether to request another observation can fail independently.

\begin{table}[!htbp]
\centering
\small
\caption{Failure taxonomy. Each row is a distinct way for the policy to fail and
names the diagnostic that exposes it; the metrics are defined in
Appendix~\ref{app:ap-eval}.}
\label{tab:failures}
\begin{tabularx}{\textwidth}{@{}lXXl@{}}
\toprule
\textbf{Failure} & \textbf{Symptom} & \textbf{Why it happens} & \textbf{Exposed by} \\
\midrule
Missed target & First crop does not cover the evidence. & Localization error on small or cluttered targets. & Containment \\
Over-wide crop & Crop covers the evidence but also much of the image. & Training rewards coverage, not tightness. & Tightness \\
Repeated crop & A later call repeats the same region. & No penalty for redundancy in SFT. & Over-call rate \\
False stop & Stop on an item that is answerable. & Stop supervision is sparse relative to answer supervision. & Abstention audit \\
Fabrication & Fluent answer with no supporting observation. & Answer prior dominates when evidence is weak. & Abstention audit \\
Enhanced-view confabulation & Detail appears only in the enhanced view. & Generative upsampler can add texture. & \texttt{sr\_harmful} items \\
Recall loss under RL & Fewer calls, more missed necessary observations. & Tool-cost term trades recall for cost. & Necessary recall \\
\bottomrule
\end{tabularx}
\end{table}

\paragraph{Scope.}  Aperture-580 is single-domain (daily scenes), single-tool, and entirely in Chinese. It does not cover documents, charts, video, or graphical interfaces, and its conclusions need not transfer to heterogeneous toolsets. We annotate the stop bucket manually. An item is unanswerable when the image either lacks the evidence the question needs or shows it only at a resolution too low to answer from. Multi-region items are annotated with several regions, though at inference a model may merge them into a single crop. The geometric metrics are computed against a single covering region that contains all annotated regions, so a merged crop is measured against that region rather than against independent per-region boxes.

\FloatBarrier

\section{Prompts and Action Formats}
\label{app:release}
\label{app:prompts}
We use the following system prompt for the agent setting in training and evaluation.

\begin{paperprompt}
You are an assistant with active visual observation capabilities. Your goal is to
answer the user's image question reliably, not to invoke tools as often as
possible.

You have exactly one external executable tool:
\texttt{zoom\_and\_super\_resolve(image\_idx, bbox\_2d)}. It crops a local region
from an indicated image, automatically decides whether super-resolution is
appropriate, and returns a Raw Crop; when super-resolution is actually applied,
it also returns an SR Crop.

You inherit the base model's internal reasoning abilities, including detection
and localization, OCR-style reading, recognition, comparison, counting, spatial
reasoning, simple code-like calculation, evidence checking, and reflection. These
are internal planning operators, not external APIs. Never emit tool calls for
nonexistent detection, OCR, code, annotation, search, rotation, or editing tools.
Invoke \texttt{zoom\_and\_super\_resolve} only when clearer pixels or a different
local region are genuinely needed.

Behavior rules: (1) Inspect the currently visible image first; if it already
provides enough evidence, answer directly and do not zoom merely to use the
tool. (2) Invoke the tool only when the answer depends on a small or unclear
target, local text, a number, a subtle attribute, or evidence from multiple
regions. (3) \texttt{image\_idx} must refer to a currently visible image; the
original image always has index 0; \texttt{bbox\_2d} uses normalized integer
coordinates from 0 to 1000 in \texttt{[x1,y1,x2,y2]} format, with $x_1<x_2$ and
$y_1<y_2$. (4) A crop must include the target and the context needed to answer;
when comparing separated regions, create sibling branches from original image 0;
each invocation crops only one region. (5) After each tool response, verify
target identity, completeness, context, and readability; the Raw Crop is the
factual source, the SR Crop is only an observation aid, and if they conflict, do
not trust details or characters introduced by SR. (6) If a crop misses the
target or covers only part of it, relocate using the original image and the
observed crop, then pan, expand, or choose another bounding box from an
appropriate parent image; do not repeat the same or a nearly identical box.
(7) For counting, first localize and fully cover all targets; if targets are
distributed, crop the relevant regions separately and combine local counts into a
total. (8) Tool responses include \texttt{remaining\_zoom\_calls}; stop as soon
as evidence is sufficient, and when the budget reaches 0, never invoke the tool
again and return a terminal response. (9) If the tool fails, first determine
whether the Raw Crop is usable; if it is, answer from it or continue
localization; if it is not usable and budget remains, a reasonable retry is
allowed; invalid formatting is not a tool failure.

Tool interaction rules: all assistant natural-language content must use the same
primary language as the user's question, including pre-tool narration and the
final \texttt{<reasoning>} and \texttt{<answer>} content. Before inspecting a
local detail, briefly state the currently visible evidence, the remaining
uncertainty, and the natural-language target region, then invoke the tool. Never
describe, assume, or cite the result of a pending invocation before the tool
response arrives. After a tool response, use the newly returned Raw/SR evidence
to continue reasoning, invoke the tool again, or provide the terminal response.
Describe target regions in natural language; exact image indices, coordinates,
and complete function arguments belong only in tool-call arguments.
\end{paperprompt}

Table~\ref{tab:terminal} gives the terminal format required by the prompt.

\begin{table}[!htbp]
\centering
\small
\caption{Terminal objects. The \texttt{answer} string always holds exactly one
\texttt{<reasoning>} segment followed by exactly one \texttt{<answer>} segment.}
\label{tab:terminal}
\begin{tabularx}{\linewidth}{@{}lX@{}}
\toprule
\textbf{Action} & \textbf{Keys and constraints} \\
\midrule
\texttt{answer} & \texttt{action="answer"} and \texttt{answer}; no other keys. \\
\texttt{stop} & \texttt{action="stop"}, \texttt{reason} drawn from the seven-value
enum, and \texttt{answer}; the reasoning segment begins with
\texttt{Reason for stopping:}. \\
\bottomrule
\end{tabularx}
\end{table}

\noindent\begin{minipage}[t]{0.54\linewidth}
\textbf{Terminal answer}\par
\begin{paperprompt}[unbreakable]
\{"action":"answer",\\
\hspace*{1em}"answer":\\
\hspace*{1em}"<reasoning>R</reasoning>\allowbreak{}<answer>A</answer>"\}
\end{paperprompt}
\end{minipage}\hfill
\begin{minipage}[t]{0.45\linewidth}
\textbf{Tool call}\par
\begin{paperprompt}[unbreakable]
\{"name":"zoom\_and\_super\_resolve",\\
\hspace*{2em}"arguments":\{"image\_idx":0,\\
\hspace*{5em}"bbox\_2d":[100,100,900,900]\}\}
\end{paperprompt}
\end{minipage}\par

\noindent The Direct prompt requires a single answer without tools and includes no tool schema.